\documentclass[a4paper, 11pt]{article}

\usepackage[english]{babel}
\usepackage[utf8]{inputenc}
\usepackage{graphicx}
\usepackage[colorinlistoftodos]{todonotes}

\usepackage{amsmath,amssymb}
\usepackage{bm}
\usepackage{dsfont}
\usepackage{algorithm,algorithmic}

\usepackage{color}

\usepackage{authblk}

\usepackage[a4paper, margin={.15\paperwidth,.08\paperheight},
marginratio=1:1]{geometry}

\definecolor{darkblue}{rgb}{0.0,0.0,0.7}

\RequirePackage[%
colorlinks = true,%
linkcolor = darkblue,%
citecolor = darkblue,%
urlcolor = darkblue, %
]{hyperref}%

\title{Mean-field inference of Hawkes point processes}

\newcommand{\ind}[1]{{\mathds{1}}_{#1}}

\newcommand{\id}{\mathbb{I}}
\newcommand{\dd}{{\rm d}}
\newcommand{\prob}{\mathbb{P}}
\newcommand{\expect}{\mathbb{E}}
\newcommand{\cov}{\mathbb{C}}
\newcommand{\var}{\mathbb{V}}

\newcommand{\logl}{\log{\mathcal L}}
\newcommand{\tr}{\mathrm{tr}}

\begin{document}

\author[1]{Emmanuel Bacry}
\author[1]{Stéphane Gaïffas}
\author[1,2]{Iacopo Mastromatteo}
\author[1,3]{Jean-François Muzy}

\affil[1]{\small Centre de Mathématiques Appliquées, \'Ecole Polytechnique and CNRS \authorcr UMR 7641, 91128 Palaiseau, France}
\affil[2]{\small Capital Fund Management, 21-23 Rue de l'Universit\'e, 75007 Paris, France}
\affil[3]{\small Laboratoire Sciences Pour l'Environnement, CNRS, Université de Corse, \authorcr UMR 6134, 20250 Corté, France}

\date{}


\maketitle

\begin{abstract}
We propose a fast and efficient estimation method that is able to accurately recover the parameters of a $d$-dimensional Hawkes point-process from a set of observations.
We exploit a {\em mean-field} approximation that is valid when the fluctuations of the stochastic intensity are small. We show that 
this is notably the case in situations when interactions are sufficiently weak, when the dimension of the system is high or when the fluctuations are self-averaging due to the large number of past events they involve.
In such a regime the estimation of a Hawkes process can be mapped on a least-squares problem for which we provide an analytic solution. 
Though this estimator is biased, we show that its precision can be comparable to the one of the Maximum Likelihood Estimator while its computation speed is shown to be improved considerably.
We give a theoretical control on the accuracy of our new approach 
and illustrate its efficiency using synthetic datasets, in order to assess the statistical estimation error of the parameters.
\end{abstract}

\section{Introduction}
The use of point processes, in particular Hawkes processes~\cite{Hawkes:1971lc,Hawkes:1971nq} is ubiquitous in many fields of applications. Such applications include, among others, geophysics~\cite{ogata1998space}, high frequency finance~\cite{Bauwens:2006,Bowsher:2007aa,review_hawkes_finance}, neuroscience~\cite{reynaud2013inference}, predictive policing~\cite{mohler2011self} and social networks  dynamics~\cite{blundell2012modelling,farajtabar2014shaping,linderman2014discovering,iwata2013discovering,zhou2013learning,trannetcodec,farajtabar2015coevolve}.
A possible explanation for the success of this model is certainly its simplicity yet ability to account for several real-world phenomena, such as self-excitement, coexistence of exogeneous and endogeneous factors
or power-law distributions~\cite{crane2008robust}.

Various estimation methods of the Hawkes process parameters have been proposed in both parametric 
and non-parametric situations. The most commonly used is to maximize the likelihood function
that can be written explicitely \cite{ogata_linear_1982} while alternative linear 
methods such as the contrast function minimization \cite{reynaud2013inference}, spectral methods \cite{bacry2012non} or estimation through the resolution of a Wiener-Hopf system \cite{bacry2014hawkes,bacry2014second} have been proposed.

In many of the above mentioned applications, especially in the field of network activities, viral propagation or 
community detection one has to handle systems of very large dimensions for which these estimation methods
can be heavy to implement.
Motivated by the goal to devise a fast and simple estimation procedure, 
we introduce an alternative approach that is inspired by
recent results \cite{delattre2014high} justifying a {\em mean-field} approximation of a Hawkes process that
supposes small fluctuations of the stochastic intensities with respect to their mean values.
This allows us to solve the estimation problem by replacing the original objective (log-likelihood) by a quadratic problem, which is much easier to optimize (see Sec.~\ref{sec:inference} below).
Note that this quadratic problem differs from the usual least-squares objective for counting processes~\cite{reynaud2013inference} (see Sec.~\ref{sec:complexity} below for a precise comparison).

We show that in a wide number of situations this new framework allows a much faster estimation
without inducing losses in precision. Indeed, we show that its bias can be
negligible and its accuracy as good as the maximum likelihood provided 
the level of endogeneity of the process is sufficiently weak 
or the interactions are suffciently ``self-averaging'' meaning that they involve a large number of events over 
past times or over components of the system (i.e., in the
large dimensional setting). Besides theoretical arguments, we give numerical illustrations of the fact that 
this our approach leads to an improvement (that can be of several orders of magnitude) 
of the classical ones based on state-of-the-art convex solvers for the log-likelihood. 

%
%


The organization of the paper is the following: in Sec.~\ref{sec:hawkes} we formally introduce the Hawkes process
and define the main notations of the paper. The mean-field inference method is defined In Sec.~\ref{sec:inference}.
We show how this method is naturally obtained using a Baysian approach in a regime where the fluctuations
of the stochastic intensity are very small.
We conduct a theoretical analysis of the domain of validity of the method by comparing its accuracy to the Maximum Likelihood Estimation. This notably allows us to provide a quantitative measure of what
weakly endogenous or ``self-averaging'' means for the interactions.
In Sec.~\ref{sec:complexity} we describe the implementation of the algorithm and compare it with other state-of-the-art algorithms.
Sec.~\ref{sec:simulation} shows the numerical results that we obtain with this method on synthetic data. 
Sec.~\ref{sec:conclusions} is devoted to concluding remarks and 
to the discussion of the possible extensions of our method,
particularly as far as penalization issues are concerned. 
The more technical parts of the discussion are relegated to the appendices.

\section{The Hawkes process}
\label{sec:hawkes}

\subsection{Definition}

Let us consider a network of $d$ nodes, in which one observes a set of \emph{events} encoded as a sequence $\{(t_m,u_m)\}_{m=1}^n$, where $t_m$ labels the time of the event number $m$ and $u_m\in \{ 1,\dots,d\}$ denotes its corresponding node. Then we can define a set of counting functions $\bm N_t=[ N^1_t,\dots,N^d_t]^\top$ as $N^i_t = \sum_{m=1}^n \delta^{i u_m} \ind{t_m \leq t}$, where $\delta^{ij}$ indicates the Kronecker delta.
These counting functions can be associated to a vector of stochastic intensities $\bm \lambda_t = [\lambda^1_t,\dots,\lambda^d_t]^\top $ defined as
\begin{equation}
    \label{eq:lambda}
    \lambda^i_t = \lim_{\dd t \to 0} \frac{\prob\left( N^i_{t+\dd t} - N^i_t = 1 \Big| \mathcal{F}_t \right)}{\dd t} \,
\end{equation}
where the filtration $\mathcal{F}_t$ encodes the information available up to time $t$.
Then the process $\bm N_t$ is called a Hawkes process if the stochastic intensities can be written as
\begin{equation}
    \label{eq:hawkes}
    \lambda_t^i = \mu^i + \sum_{j=1}^d \int_{0}^t \dd N^j_{t'} \Phi^{ij}(t-t') \, ,
\end{equation}
where $\bm{\Phi}(t-t')=[ \Phi^{ij}(t-t') ]_{1\leq i,j\leq d}$ is a \emph{component-wise positive}, \emph{causal} (i.e., whose support is in $\mathbb{R}^+$), \emph{locally $L^1$-integrable} matrix kernel representing (after proper normalization) the probability for an event of type $j$ occurring at time $t'$ to trigger an event of type $i$ at time $t$, while $\bm \mu = [\mu^i]_{i=1}^d$ is a vector of \emph{positive} exogenous intensities (see Ref.~\cite{daley1988introduction} for a more rigorous definition).
It is well known that a sufficient condition for the intensity processes $\lambda^i_t$ to be stationnary (i.e., for the processes $N^i_t$ to have stationnary increments) is the so-called {\bf Stability Condition}
\begin{equation}
\label{eq:sc}
{\bf (SC) :}~~~~~ ||{\bm \Phi}||_1 < 1
\end{equation}
where $||.||_1$ stands for the spectral norm of the $d \times d$ matrix $\left[\int_0^{+\infty}\Phi^{ij}(t)\dd t\right]_{1\le i,j \le d}$ made of the $L^1$ norms of each kernel $\Phi^{ij}(t)$.

In the following discussion, we will restrict our attention to stable Hawkes processes (in the sense of {\bf (SC)}) for which the matrix $\bm{\Phi}(t)$ can be written as
\begin{equation}
    \label{eq:hawkes_parametrization}
    \Phi^{ij}(t) = \sum_{q=1}^p \alpha^{ij}_q g_q(t)
\end{equation}
where we have introduced a set of $p > 0$ known \emph{basis kernels} $g_q(t)$ satisfying $$\int_0^\infty \dd t \, g_q(t) =1$$.


\subsection{Notations}
For the sake of conciseness, it will be useful to preliminary introduce some notation.
\begin{itemize}
    \item We will use the integer indexes $a \in \{1, \ldots, pd\}$ in order to identify pairs $(j_a,q_a) \in \{1,\dots , d\} \times \{ 1,\dots p \}$. Consequently, summing over $a$ allows to run both over the $d$ components and the $p$ basis functions of the kernels. While the set of indices $i,j,k \dots$ will be employed to label single nodes, the notation $a,b,c,\dots$ will be used to label a pair (node/basis kernel).
    In that respect we set 
    \begin{itemize}
    \item $\alpha^{ia} \equiv \alpha^{i j_a}_{q_a} $,
    \item $g^{a}(t-t') \equiv g_{q_a}(t-t')$ and
    \item $\dd N^a_t \equiv \dd N^{j_a}_t$.
    \end{itemize}
    \item For convenience we will also define the deterministic process $N^0_t = t$, which will be associated with the kernel $g^{0}(t-t')=\delta(t - t' - \dd t)$ equal to a Dirac delta function shifted by an infinitesimal amount $\dd t > 0$.

    \item According to previous notations, Eq.~(\ref{eq:hawkes}) can be compactly rewritten as
    \begin{equation}
        \label{eq:hawkes_compact}
        \lambda_t^i = \sum_{a=0}^{dp} \theta^{ia} \int_0^t \dd N^a_{t'} g^{a}(t-t')\,
    \end{equation}
    where the parameters $\theta^{ia}$ refer to the Hawkes parameters, namely\footnote{Note that, in order for the parameters $\theta^{ia}$ to be dimensionally homogeneous, we need to assume a dimensionless $\mu$, while $g^{0}(t-t')$ is taken to be of dimension $[t]^{-1}$.
    Moreover, we implicitly assume $N^0_t = \omega t = t$ to be dimensionless through a suitable choice of a unit
    (i.e, we take $\omega=1$).}
    $$\theta^{i0} = \mu^i \mbox{~~and } ~~\theta^{ia} = \alpha^{ia},~~\forall a>0.$$ 
    \item We will adopt the notation $x$ for scalars , ${\bm y} = [y^{a}]_{0\leq a\leq dp}$ for vectors  and $\bm{\mathsf{Z}} = [Z^{ab}]_{0 \leq a,b\leq dp}$ for square matrices. Correspondingly, we define the usual matrix-vector composition by $\bm{\mathsf{Z}} \bm y = [\sum_{b=0}^{dp} Z^{ab} y^{b}]_{0\leq a\leq dp}$ and the matrix-matrix product as
    $\bm{\mathsf{W}} \bm{\mathsf{Z}} = [\sum_{b=0}^{dp} W^{ab} Z^{bc}]_{0\leq a,c\leq dp}$.
    
    \item If ${\bm y}$ is a vector $||{\bm y}||$ will refer to the $L^2$ norm of ${\bm y}$ whereas 
    if $\bm{\mathsf{Z}}$ is a matrix, $||\bm{\mathsf{Z}}||$ will refer to its spectral norm. Moreover if ${\bm \Phi}(t)=[\Phi^{ij}(t)]_{1\le i,j \le d}$ is a matrix of functions,
    $||{\bm \Phi}||_1$ will refer to the spectral norm of the matrix $[\int_0^{+\infty}\dd t\, |\Phi^{ij}(t)|]_{1\le i,j \le d}$.

    \item Hereafter, we will have to handle collections of $d$ scalars, vectors or matrices indexed by $i\in\{1,\dots,d\}$, that we will write as 
    $x^i$ (for scalars), ${\bm y}^i = [y^{ia}]_{0\leq a\leq dp}$ (for vectors) and $\bm{\mathsf{Z}}^i = [Z^{iab}]_{0\leq a,b\leq dp}$ (for matrices).
\end{itemize}
According to these conventions,
The goal of this paper is to present a mean-field framework for estimation of 
the $d$ vectors of parameters $\bm{\theta}^i = [\theta^{ia}]_{0\leq a\leq dp} $ given a set of observations $N_t$. In the following the set of all the $\bm\theta^i$ parameters will be often referred to as ${\bm \theta} = \{\bm \theta^i\}_{1 \le i \le d}$.

\subsection{The Likelihood function}
The probability for a Hawkes process parametrized by $\bm\theta$ to generate a trajectory $N_t$ in a time interval $t \in [0,T]$ has been first computed in Ref.~\cite{ogata_linear_1982}, and is given by 
\begin{equation}
    \label{eq:likelihood}
    \prob(\bm N_t | {\bm \theta}) = e^{- \sum_{i=1}^d \int_0^T \dd t \, \lambda^i_t } \, \prod_{m=1}^n \lambda^{u_m}_{t_m} \,
\end{equation}
which implicitly depends on ${\bm \theta}$ through the stochastic intensities $\lambda^i_t$.
We define the negative log-likelihood
\begin{equation}
\label{eq:ll}
\mathcal{L}(\bm N_t , {\bm \theta}) = -\log\prob(\bm N_t|{\bm\theta}).
\end{equation}    
Note that the maximum-likelihood estimator (that we will compare to our mean-field estimator to throughout this paper)
\begin{equation}
    \label{eq:MLE}
    {\bm \theta_{MLE}} = \textrm{argmin}_{{\bm \theta}} \mathcal{L}(\bm N_t , {\bm \theta}),
\end{equation}
is the most commonly used estimator employed in the literature \cite{ogata_linear_1982}.

\section{Inference and mean-field approximation}
\label{sec:inference}
\subsection{A Bayesian approach}
From a Bayesian standpoint, the probability for an observed sample $\bm N_t$ to be generated by a Hawkes process parametrized by $\bm\theta$ is given by the \emph{posterior distribution}
\begin{equation}
    \label{eq:bayes}
    \prob(\bm \theta| \bm N_t) = \frac{\prob(\bm N_t | \bm \theta)\prob_0(\bm \theta)}{\prob( N_t )} \,
\end{equation}
where $\prob_0(\bm \theta)$ is a \emph{prior} which we assume to be flat (see Sec.~\ref{sec:conclusions} below for other choices). Hence, we will be interested in averaging the inferred couplings over the posterior, so to compute their averages and their covariances
\begin{eqnarray}
    \label{eq:av_theta}
    \expect \theta^{ia} &=& \int \dd{\bm\theta} \, \prob(\bm \theta | \bm N_t ) \, \theta^{ia} \\
    \label{eq:fl_theta}
    \cov(\theta^{ia},\theta^{jb}) &=& \int \dd{\bm\theta} \,  \prob(\bm \theta | \bm N_t ) \theta^{ia}\theta^{ib} - \expect\theta^{ia}\expect\theta^{jb} \, , \nonumber \\
    &&
\end{eqnarray}
where we are writing $\int \dd{\bm\theta} = \int_{\mathbb{R}_+^{d(dp+1)}} \prod_{1\leq i\leq d,0\leq a \leq dp} \dd\theta^{ia}$.
This can be technically done by introducing the \emph{partition function} $Z({\bm s})$ defined as the Laplace transform of the posterior
\begin{equation}
    \label{eq:part_func}
    Z({\bm s}) = \int \dd{\bm\theta} \,  \prob(\bm \theta | \bm N_t ) e^{-T \, \sum_{i=1}^d {\bm s}^{i\top} {\bm\theta}^i  }
\end{equation}
where ${\bm s} = \{ {\bm s}^i \}_{1\le i \le d}$ is a collection of $d$ Laplace-parameter vectors ${\bm s}^i = [s^{ia}]_{0\leq a \leq dp}$.
Its computation allows one to obtain Eqs.~(\ref{eq:av_theta}) and~(\ref{eq:fl_theta}) by differentiating the \emph{free-energy density} $f({\bm s})= - \log Z({\bm s})/T$, so that
\begin{eqnarray}
    \expect\theta^{ia} &=& \partial_{s^{ia}} f(\bm s) |_{\bm s = 0} \\
    \cov(\theta^{ia},\theta^{jb}) &=& - \frac{1}{T}\partial_{s^{ia}}\partial_{s^{jb}} f(\bm s) |_{\bm s = 0} 
\end{eqnarray}
We will be interested in finding an approximated expression for the free-energy density $f({\bm s})$ allowing to compute efficiently Eqs.~(\ref{eq:av_theta}) and~(\ref{eq:fl_theta}). 

\vskip .2cm
\noindent
{\bf Remark: } Let us point out that, since we are interested in derivatives of the free energy, we can drop additive $\bm s$-independent terms in its computation. 
This will be done implicitely all along the computation.
\vskip .2cm

\noindent
Let us first use Eq.~(\ref{eq:likelihood}) so to write explicitly the partition function as
\begin{equation}
    \label{eq:part_func_expl}
    Z({\bm s}) = \int \dd{\bm\theta} \,  e^{-T \,\sum_{i=1}^d({\bm s}^i + {\bm h})^\top {\bm \theta}^i + \sum_{m=1}^n \log \lambda^{u_m}_{t_m}}
\end{equation}
where we have dropped multiplicative $\bm s$-independent terms, and introduced auxiliary vector $\bm h = [h^{a}]_{0\leq a \leq dp}$ equal to
\begin{equation}
    h^{a} =  \frac{1}{T}\int_0^T \dd t \int_{0}^t \dd N_{t'}^a g^{a}(t-t') \, .
\end{equation}
Note that in the limit of large sample size $T\to\infty$ and in the vicinity of $\bm s = 0$, the integral is dominated by the maximum of the exponential corresponding to the maximum likelihood estimator ${\bm \theta_{MLE}}$ \eqref{eq:MLE} which is commonly employed in the literature relating to parametric inference of Hawkes processes~\cite{ogata_linear_1982}.

Let us also point out that $Z({\bm s})$ can be rewritten as:
\begin{equation}
\label{eq:part_func_expl2}
Z({\bm s})
=  \int \dd{\bm\theta} \,  e^{\sum_{i=1}^d \left(-T({\bm s}^i + {\bm h})^\top {\bm \theta}^i + \int_0^T dN_t^i \log \lambda^{i}_{t}\right)}
= \prod_{i=1}^d \int \dd{\bm\theta}^i   e^{-T ({\bm s}^i + {\bm h})^\top {\bm \theta}^i + \int_0^T dN_t^i \log \lambda^{i}_{t}}
\end{equation}
and therefore the partition function is a product of independent partition functions associated with 
each node $i\in \{1,\ldots,d\}$. This means that the inference problem factorizes in single node problems associated with each vector ${\bm \theta}^i$.
    
\subsection{Mean-field approximation} 
Our strategy in order to approximate Eq.~(\ref{eq:part_func_expl}) is to suppose the fluctuations of the $\lambda_t^i$ to be small with respect to their empirical averages 
\begin{equation}
\label{eq:defLambda}
\bar {\bm \Lambda} = \{ \bar  \Lambda^i \}_{1\le i\le d} = \left\{ \frac{N^i_T}{T} \right\}_{1\le i\le d} \, .
\end{equation}
This allows us to approximate the partition function appearing in Eq.~(\ref{eq:part_func_expl}) by a quadratic expansion of the logarithmic functions around the empirical averages $\bar{\bm\Lambda}$.
Let us first suppose that the Hawkes process is stable in the sense of {\bf (SC)} (see \eqref{eq:sc}). We have seen that it means that the processes $[\lambda_t^i]_{1\leq i\leq d}$ are stationary. The 
mean-field expansion basically assumes that each process $\lambda_t^i$ has small fluctuations around its 
empirical averaged value $\bar \Lambda^i$. More precisely, let us suppose that the following  {\bf Mean-Field Hypothesis} 
holds
\begin{equation}
\label{eq:mfas1}
{\mbox {\bf  (MFH) :}}~~~~~~  r_i =  \frac{|\lambda^i_t-\bar \Lambda^i|}{\bar \Lambda^i} \ll 1,
\end{equation}
then
\begin{equation}
  \label{eq:log_lambda}
  \log \lambda^i_t \simeq \log \bar \Lambda^i + \frac{\lambda^i_t-\bar \Lambda^i}{\bar \Lambda^{i}}-\frac{(\lambda^i_t-\bar \Lambda^i)^2}{2 (\bar \Lambda^{i})^2} \; .
\end{equation}
In order to get $Z(\bm s)$ (see \eqref{eq:part_func_expl2}), we need to compute $\int_0^T dN_t^i \log \lambda^{i}_{t}$ for all $i=1,\ldots,d$. Substituing \eqref{eq:log_lambda} in this expression leads to three terms. 
\begin{itemize}
\item The first term 
$$\int_0^T dN_t^i \log \bar \Lambda^i$$ 
is independent of both ${\bm \theta}$ and ${\bm s}$, thus it can be dropped.\footnote{Had we introduced $\bar \Lambda$ as a variational parameter to estimate self-consistently, this term would have contributed to fixing it to its optimal value. This is non-necessary in this case, since we are implicitly forcing the empirical value of $\bar\Lambda$ to be the mean-field parameter.}
\item The second term 
$$\int_0^T dN_t^i \log  \frac{\lambda^i_t-\bar \Lambda^i}{\bar \Lambda^{i}}$$
after dropping ${\bm \theta}$ and ${\bm s}$-independent terms lead to
$$T \sum_{a=0}^{dp}  \theta^{ia} k^{ia}$$
where we have introduced a collection of $d$ vectors $\bm k^i = [k^{ia}]_{0\leq a \leq dp}$ defined as:
\begin{equation}
\label{eq:aux_k}
k^{ia}   =  \frac{1}{N^i_T} \int_{t\neq t'} \dd N^i_t \dd N^a_{t'} g^{a}(t-t') 
\end{equation}
\item The third term (after straightforward computations, dropping again ${\bm \theta}$ and ${\bm s}$-independent terms) leads to 
$$
- \frac T {2}
 \sum_{a,b=0}^{dp} \theta^{ia}\theta^{ib} J^{iab} 
  +  T  \sum_{a=0}^{dp}   \theta^{ia} k^{ia},
$$
where we have introduced a collection of $d$ matrices $\bm{\mathsf{J}}^i = [J^{iab}]_{0\leq a,b \leq dp}$ defined as:
\begin{equation}
\label{eq:aux_j}
J^{iab}   =  \frac{T}{(N^i_T)^2} \int_t \int_{t'\neq t} \int_{t''\neq t} \dd N^i_t \dd N^a_{t'} \dd N^b_{t''} \\ 
 g^{a}(t-t') g^{b}(t-t'') \, .
\end{equation}
\end{itemize}
It then results from \eqref{eq:part_func_expl2} that, up to a constant factor,
\begin{equation}
    \label{eq:free_en_approx}
    Z(\bm s) \simeq \int \dd \bm\theta \, e^{
    -T \sum_{i=1}^d
    \left(   ({\bm s}^i + {\bm h} - 2 {\bm k}^i)^\top {\bm \theta}^i
        + \frac{1}{2} \bm{\theta}^{i\top} \bm{\mathsf{J}}^i \bm{\theta}^i
    \right)}.
\end{equation}
The integral appearing in Eq.~(\ref{eq:free_en_approx}) can be evaluated analytically as a simple
Gaussian integral\footnote{As the region of integration in $\dd \bm \theta$ is the positive orthant $\mathbb{R}^{d(dp+1)}_+$, Eq.~(\ref{eq:mf_free_en}) should in principle include extra terms preventing the emergence of negative couplings. We prefer to extend the region of integration to the whole space $\mathbb{R}^{d(dp+1)}$, so to include potentially negative couplings. A saddle-point expansion of Eq.~(\ref{eq:free_en_approx}) reveals indeed that these differences are subleading in $T$.}, so that one can write the free-energy density as
\begin{equation}
    \label{eq:mf_free_en}
    f(\bm s) = - \frac{1}{2} \sum_{i=1}^d \left[ (2{\bm k}^i - {\bm h}^i - {\bm s}^i)^\top \bm{\mathsf{C}}^i (2{\bm k}^i - {\bm h}^i - {\bm s}^i) \right] \, ,
\end{equation}
where $\bm{\mathsf{C}}^i$ denotes the inverse of matrix $\bm{\mathsf{J}}^i$. 
Under this approximation, one obtains
\begin{eqnarray}
    \label{eq:estimator_theta}
    \expect\theta^{ia} &\simeq& \sum_{b=0}^{dp} C^{iab} (2 k^{ib} - h^{b}) \equiv \theta^{ia}_{MF}\\
    \label{eq:covariance_theta}
    \cov(\theta^{ia},\theta^{jb}) &\simeq& \frac{\delta^{ij} C^{iab}}{T} \, .
\end{eqnarray}
These are the central equations of our paper: {\bf Eq.~(\ref{eq:estimator_theta}) expresses the definition of our mean-field estimator}
while Eq.~(\ref{eq:covariance_theta}) expresses the convergence rate to the expected value of the estimator, so that $\bm \theta - \expect\bm \theta \sim T^{-1/2}$.

\subsection{Validity conditions for the mean-field approximation} 
\label{sec:3.3}
As discussed previously, the mean-field approximation is likely to be 
pertinent in the regime described by {\bf (MFH)} \eqref{eq:mfas1}, i.e., 
when the $\bm \lambda_t$ calculated under the inferred parameters have
small fluctuations with respect to the empirical averages $\bar {\bm \Lambda}$.
However the computation we performed in the previous section 
(relying on a second order truncation of $\log \lambda^i_t$) can be hardly used to obtain
a control on the accuracy of this approximation or to define its domain of validity.
These problems are addressed in Appendix \ref{app:mle_versus_mf_solution} where we show that 
the error of the mean-field estimates ${\bm \theta}^i_{MF}$ with respect to the maximum-likehood estimates
${\bm \theta}^{i}_{MLE}$ \eqref{eq:MLE}
\begin{equation}
 \bm{\delta \theta}^{i} = {\bm  \theta}^i_{MF} - {\bm \theta}^{i}_{MLE}
\end{equation} 
is directly bounded by the ratio of the empirical variance of $\lambda^i_t$ to its empirical mean. In particular, at leading order in the fluctuations and in the limit of large $T$, one has
\begin{equation}
\label{eq:variance_bound1}
	||\bm{\delta \theta}^{i}|| \leq \frac{\bar \var(\lambda^i_t)}{(\bar \Lambda^i)^2} ||\bm{\mathsf{C}}^i|| ||\bm{\bar \Lambda}||
\end{equation}
where $||.||$ refers to the $L^2$ norm in case of vectors and spectral norm in case of matrices, $\bar{\bm \Lambda} = [\bar \Lambda^a]_{a=0}^{dp}$ and
where the variance corresponds to the \emph{empirical} one computed under the saddle-point parameters $\bm{\theta}_{MLE}$.
We see that when the {\em fluctuation ratio} $\bar \var({\lambda^i_t})/(\bar \Lambda^i)^2$ is very small, 
the results of the mean-field estimate can be very close to the maximum-likelihood estimate.

\vskip .3cm
\paragraph{Analysis of the fluctuation ratio}
The natural question that follows these considerations concerns the characterization of the 
situations when this fluctuation ratio is very small. 
In appendix \ref{sec:homogeneous_system}, we establish these conditions in the particular case of a perfectly homogeneous 
Hawkes process. We show that in the limit of small $\bm ||\bm\Phi||_1$ (Weak endogeneity case (i) below), or in the limit of large dimension $d$ or slow interactions (Self-averaging interactions case (ii) below) the fluctuation ratio is small. More precisely (see Eq. \eqref{eq:fluc}), we find that it is directly controlled by the spectral norm $||\bm\Phi||_1$, the system dimension $d$ and the interaction characteristic time-scale $\tau_g$ (defined in Appendix \ref{sec:homogeneous_system}):
\begin{equation}
 \label{eq:fratio}
 \frac{\var(\lambda^i_t)}{(\Lambda^i)^2}  \sim \frac 1 {\Lambda^i \tau_g} \left( \frac{||\bm\Phi||^2_1}{d(1-||\bm\Phi||_1)^{2-1/\eta}} \right).
\end{equation}
Let us recall that in the exponential kernel case, one has $\eta = 1$ (see Appendix \ref{sec:homogeneous_system}).
Intuitively, this characterization indicates that the mean-field approximation is valid whenever one of these conditions is met:
\begin{itemize}
    \item[(i)] {\em Weak endogeneity} ($|| \bm\Phi ||_1 \ll 1$ ): If the spectral norm $|| \bm\Phi ||_1$ (that controls the level of endogeneity of the Hawkes process) is much smaller than one, then the intensity vector $\lambda_t$ is dominated by the exogenous component, and one can expect a very small fluctuation ratio.
    
    \item[(ii)] {\em Self-averaging interactions} ($d \bar \Lambda \tau_g \gg 1$): Even if the system is not weakly endogenous (i.e. $|| \bm\Phi ||_1$
    is not necessarily small), if the component-wise intensities are determined by a very large number of events with comparable contributions, a law of large numbers leads to small fluctuation ratios.
    This can notably occur in two different situations. First in large dimensional ``quasi-homogenous'' systems where
    a large number of events associated with each of the components contribute to the intensity.
    This regime corresponds to large $d$ in Eq. \eqref{eq:fratio} and has been studied notably in Ref.~\cite{delattre2014high}.
    The second situation is the case of ``slow interactions'' when a large number of past events equally impact the intensity. If the characteristic time scale of the kernel $\bm g(t)$ is large with respect to the typical inter-event distance, then one expects the past contribution to the instantaneous intensity $\lambda_t^i$ to have small fluctuations. This regime corresponds to a large $\bar \Lambda \tau_g$ in Eq. \eqref{eq:fratio}.
\end{itemize}

\begin{figure}[htbp]
    \centering%
    \includegraphics[scale=0.8]{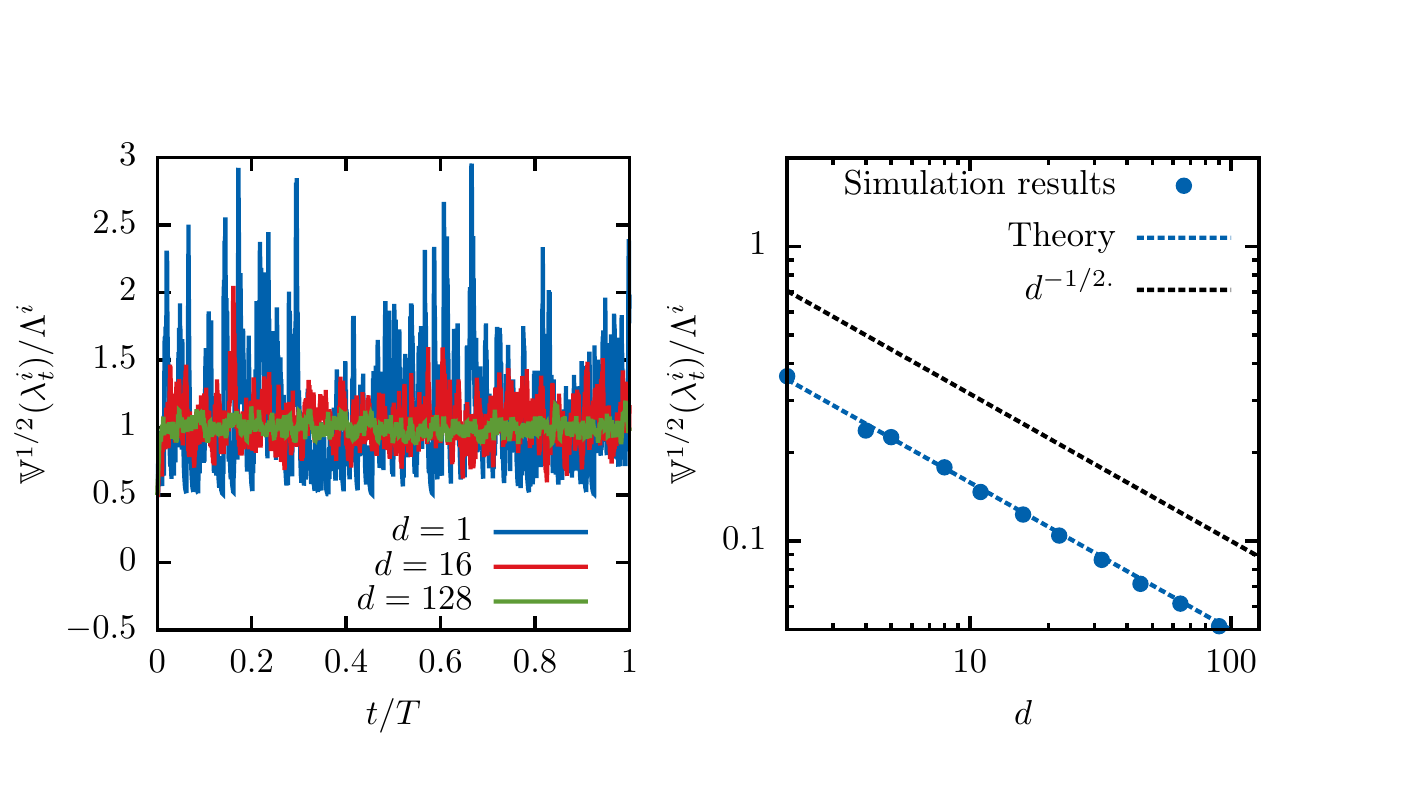}
    \caption{
    Fluctuation ratio $r_i=\bar \var^{1/2}(\lambda^i_t)/\bar \Lambda^i$ of the intensity of a simulated Hawkes process
     as a function of the number of components
    $d$ for a fixed value of $\bm\Lambda$ and the interaction parameter $||\bm\Phi||_1 = 0.5$ against the theoretical prediction of Eq.~(\ref{eq:flucexp}). By
    increasing the size of the system with a fixed value of the intensity, the fluctuations are found to decrease
    as $d^{-1/2}$. Indeed, the theoretical predictions (dashed blue line) perfectly match the results of simulation (dots). The parameters of the simulation are defined
    as in Sec.~\ref{sec:simulation}, and are equal to $\beta=\mu=1$. The
    interaction matrix has the two-block structure as illustrated in Sec.~\ref{sec:simulation}.}
    \label{fig:scaling_fluct}
\end{figure}
\begin{figure}[h]
    \begin{center}
        \includegraphics{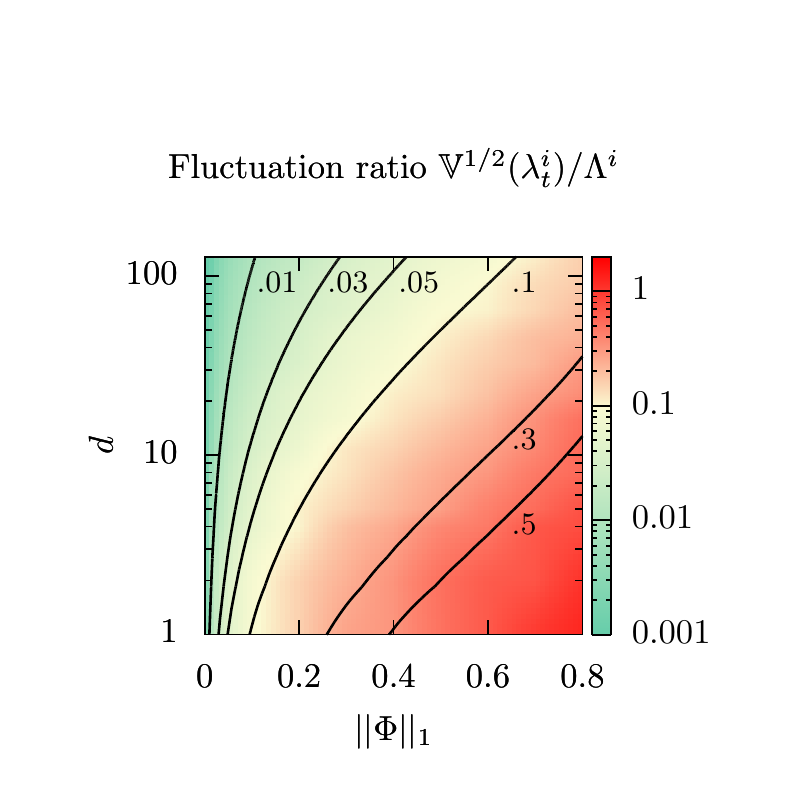}
    \end{center}
    \caption{
     Fluctuation ratio $r_i = \bar \var^{1/2}(\lambda^i_t)/\bar \Lambda^i$ for a system with variable number of components $d = \{1,\dots, 128\}$ and interaction strengths $||\bm \Phi||_1 \in [0,0.8)$, given a fixed value of the intensities $\bar {\bm\Lambda} $. We have chosen $p=1$ and $g_1(t) = \ind{t>0} \beta e^{-\beta t}$ (with $\beta = 1$) and chosen a two-block structure for the matrix $\bm \alpha_1$ with entries equal to either $\alpha^{ij}_1 = 0$ or $\alpha^{ij}_1 = ||\bm \Phi||_1 / c$, with  $ c \in \{\lceil d/2 \rceil, \lfloor d/2 \rfloor\}$, as described in Sec.~\ref{sec:simulation}. The other simulation parameters are set to $\mu = 1-||\bm{\Phi}||_1$, $\beta_1=1$. The figure shows that larger values of $||\bm \Phi||_1$ increase the size of the fluctuations. Conversely, the fluctuations can be reduced by an arbitrary amount by distributing the interaction among a large number of components $d$. The theoretical predictions of Eq.~(\ref{eq:flucexp}) correspond to the contour lines, which accurately reproduce the results of simulation.}
    \label{fig:contour_scaling_fluct}
\end{figure}

The perfect homogeneous case is a very good toy model, it has the main ingredients that make the mean-field hypothesis valid. The result we so-obtained in the Appendix \ref{sec:homogeneous_system} reproduce much more complex situations. 
Indeed, in the following we considered a
Hawkes process with a 2-block interaction matrix and exponential kernels fully specified in Sec.~\ref{sec:simulation}.
Figs~\ref{fig:scaling_fluct} and \ref{fig:contour_scaling_fluct} show that the formula for the fluctuation ratio
$r_i =  \sqrt{\frac{\var(\lambda^i_t)}{(\Lambda^i)^2}}$ (see Eq.~\eqref{eq:fratio} or equivalently \eqref{eq:fluc} for general kernel case and Eq.~\eqref{eq:flucexp} for exponential kernels) lead to a perfect prediction of the simulated curve.

The right plot of Fig.~\ref{fig:scaling_fluct} shows that the fluctuation ratio $r_1$ of the first block ($i=1$) behaves in very good agreement with Eq. \eqref{eq:flucexp}. It scales like $d^{-1/2}$ when the dimension varies.
Fig.~\ref{fig:contour_scaling_fluct} shows the same ratio 
as a function of both the spectral norm $||\bm \Phi||_1$ and the dimension $d$. Let us point out that, when varying the norm $||\bm\Phi||_1$  and/or the dimension $d$, the average expectation $\bm\Lambda$ is held fixed.
One can notably see that, both
(i) for a fixed value of the dimension, there exists a sufficiently small interaction $||\bm \Phi||_1$ such that the fluctuation
ratio can be arbitrarily small and (ii) for a fixed 
value of the norm (even large), there exists a sufficiently large system dimension $d$ such that the fluctuation
ratio can be arbitrarily small. The displayed contour plots are the ones predicted by our analytical 
considerations \eqref{eq:flucexp} proved in Appendix \ref{sec:homogeneous_system} 
(see Eqs \eqref{eq:fluc} and \eqref{eq:flucexp}). The theoretical prediction appears to be in good agreement with simulated data.

Finally, let us remark that while above considerations are useful for assessing the validity of the mean-field approximation under some specific assumption for the model, it is possible to bound \emph{a priori} the approximation error $\bm{\delta\theta}$ by exploiting the results of App.~\ref{app:cumulant_expansion}, that allow one to estimate the mean-field error by supplying the empirical cumulants of the data, without requiring any assumption about the underlying model.


\vskip .3cm
\paragraph{Estimation error and validity condition}
Since it basically amounts in solving a linear system, it is clear that the mean-field estimator will perform much faster than any MLE based estimator (see next section for deatiled analysis on the complexity). The idea is to get, while being much faster, an estimation as precise as the MLE. This is true provided $||\bm{\delta \theta}^{i}||$ is smaller
than the statistical error $||\bm {\Delta \theta}^i|| = ||\bm {\theta}_{MLE}^i -\bm {\theta}_{true}^i||$ associated with the maximum likelihood estimator.
One can obtain a self-consistent estimation of the regime in which the error is dominated by its statistical component by using Eq.~\eqref{eq:covariance_theta}):
$$
||\bm {\Delta \theta}^i|| \simeq \left( \frac{\tr (\bm{\mathsf{C}}^i)}{T} \right)^{1/2} ,
$$
that in principle is valid under the mean-field approximation. Figure~\ref{fig:error_estimation} below shows that this assumption is correct under a large choice of parameter values.
Then a simple condition for the mean-field method to perform as well as the MLE is
$$ ||\bm{\delta \theta}^{i}|| < \left( \frac{\tr (\bm{\mathsf{C}}^i)}{T} \right)^{1/2}.$$
Eq.~\eqref{eq:variance_bound1} provides a sufficient condition for this to hold (after identifying empirical averages and statistical averages), namely
$$ \frac{\var(\lambda^i_t)}{(\Lambda^i)^2} ||\bm{\mathsf{C}}^i|| ||\bm{\Lambda}|| \lesssim \left( \frac{\tr (\bm{\mathsf{C}}^i)}{T} \right)^{1/2}.$$
Eq. \eqref{eq:fratio} has been established in the perfectly homogeneous case for which
$||\bm\Lambda|| = \sqrt{d} \Lambda^i$. Moreover, we use very conservatively the inequality $||\bm{\mathsf{C}}^i|| \leq \tr (\bm{\mathsf{C}}^i)$, which is likely to be largely underestimating the statistical error, so that in practice the performance of the MF algorithm is much better than what appears from Eq.~(\ref{eq:bound_stat_err}) below (see Fig.~\ref{fig:err_vs_T}). Indeed, plugging~\eqref{eq:fratio} and assuming the scaling $||\bm{\mathsf{C}}^i||\sim (\Lambda^i)^{-1}$ in the last equation leads to the sufficient condition
$$ \frac{1}{\sqrt{\Lambda^i}\tau_g} \left( \frac{ \, ||\bm\Phi||^2_1}{d^{1/2} (1-||\bm\Phi||_1)^{2-1/\eta}} \right) \lesssim \left( \frac{1}{T} \right)^{1/2}.$$
Therefore
\begin{equation}
    \label{eq:bound_stat_err}
    T \lesssim T^\star  = \Lambda^i\tau_g^2 \left( \frac {d (1-||\bm\Phi||_1)^{4-2/\eta}} {||\bm\Phi||^4_1}\right) .
\end{equation}

Although far from being optimal, this rough bound provides a qualitative idea of how the different values of $d$, $||\bm\Phi||_1$ and $\tau_g$ can contribute in determining the performance of the mean-field approximation.
We will see in Sec.~\ref{sec:simulation}, that in practive $T$ can be very large
for moderate system dimension and values of the spectral norm.
In other words, the situations where the performance of the 
mean-field method is comparable to those of the maximum likehood are
very easy to meet, in particular for large system sizes.
In Sec.~\ref{sec:simulation} we will illustrate this point by showing that in a broad range of regimes of $(||\bm\Phi||_1,d,T)$ we will be able to exploit the condition $T < T^{\star}$, thus outperforming the maximum-likelihood algorithm through the superior numerical performance of the mean-field approximation.


\section{Implementation and complexity}
In this section we want to illustrate the numerical implementation of the mean-field algorithm, by comparing its complexity with the one of other state-of-the art algorithm.
We find that the mean-field algorithm has a cost corresponding to $dp$ iterations of a gradient ascent in the likelihood, both for the quasi-Newton maximisation of the likelihood (Sec.~\ref{sec:MLE}) and the expectation-maximisation algorithm (Sec.~\ref{sec:EM}). This is typically much faster than the times required by these methods to achive convergence (Fig.~\ref{fig:time_perf}). The complexity of the contrast function algorithm (Sec.~\ref{sec:CF}) is in principle of the same order of the mean-field approximation, but is in practice much slower (see Fig.~\ref{fig:time_perf}).
\label{sec:complexity}
 \subsection{Mean-field (MF) algorithm}
 \label{sec:mf_algo}
 The simplest procedure to use in order to obtain the estimator Eq.~(\ref{eq:estimator_theta}) and its associated covariance matrix is summarized in Algorithm~\ref{algo:mf}.

 \begin{algorithm}
 \caption{Calculation of the mean-field estimator $\expect\theta^{ia}$ and its covariance $\cov(\theta^{ia}\theta^{jb})$.}
 \label{algo:mf}
 \begin{algorithmic}[1]
 \renewcommand{\algorithmicrequire}{\textbf{Input:}}
 \renewcommand{\algorithmicensure}{\textbf{Output:}}
 \REQUIRE $\bm N_t,\bm g(t)$
 \ENSURE  $\expect\theta^{ia}$, $\cov(\theta^{ia},\theta^{jb})$
  \STATE Compute the auxiliary functions $\bm h, \bm{k}^i$ and $\bm{\mathsf J}^i$ as shown in Appendix~\ref{app:aux_func}
  \STATE Invert the $d$ matrices $\bm{\mathsf{J}}^i$
  \STATE Evaluate Eqs.~(\ref{eq:estimator_theta}) and~(\ref{eq:covariance_theta})
 \RETURN $\expect\theta^{ia}$ , $\cov(\theta^{ia},\theta^{jb})$
 \end{algorithmic} 
 \end{algorithm}

Note that the complexity of the preprocessing phase in which one computes the auxiliary functions $\bm h, \bm k^i$ and $\bm{\mathsf{J}}^i$ is dominated by the computation of the $\bm{\mathsf{J}}^i$, and scales as $O(d^2 p \Lambda T \times \max(\Lambda T, d p))$. The inversion of the matrices can be performed in a time bounded by $ O(d^4 p^3) $. In order to speed up the computation, one can use the following tricks:
\begin{itemize}
    \item The pre-processing time can be reduced by fixing a threshold $\epsilon$ so to cutoff the integrals over the basis kernels $\bm g(t)$ when $t$ is sufficiently large. This reduces the pre-processing time to $O(d^2 p \Lambda T \times \max(\Lambda t, d p))$ , where $t$ is an $\epsilon$-dependent bound on the characteristic time required for the kernels $g^{a}(t)$ to decay below the threshold $\epsilon$ (see Appendix~\ref{app:aux_func}).

    \item If one is not interested in the covariances $\cov(\theta^{ia},\theta^{ib})$, the inversion of the matrix can be avoided, by replacing it with the solution of the $d$ linear systems $\bm{\mathsf{J}}^i \, \bm{\theta}_{MF}^i = 2 \bm k^i - \bm h$. This substantially reduces the time required to find $\bm{\theta}_{MF}$ in step 2.\ and 3.\ of Algorithm~\ref{algo:mf}, as the matrix inversion can be traded with a faster linear solver (e.g., Cholesky decomposition, or
    iterative algorithms such as BFGS~\cite{liu1989limited}).

    \item The case of weak endogeneity described above allows one to perform an approximate inversion of the matrices $\bm{\mathsf{J}}^i$, as shown in Appendix~\ref{app:small_fluct}, reducing the time required for the solution of the $d$ linear systems from $O(d^4 p^3)$ to $O(d^3 p^2)$ (corresponding to $d$ matrix-vector compositions). Note indeed that this is expected to yield inaccurate results when the fluctuations of the $\bm N_t$ are of a larger order than $d^{-1}$, even though the elements of the $\bm\alpha$ matrix are individually small (see again Appendix~\ref{app:small_fluct}).
\end{itemize}
 The time of computation is typically dominated by the preprocessing phase up to very large sizes. As an illustrative example, a parallel implementation in \texttt{python} + \texttt{c} on a four-core machine with a 3.40 Hz CPU the calculation of the auxiliary functions takes $\sim 6 \times 10^2$~s for $d=128$, $p = 1$ and $n = 2 \times 10^3$ events, while the matrix inversions can be performed in a time $~10^{-1}$~s. Finally, we remark that the algorithm can be straightforwardly parallelized across components, meaning that a factor $d$ can be gained in both the preprocessing and the matrix inversion phase.

\subsection{Maximum-Likelihood Estimation (MLE)}
    \label{sec:MLE}
    This class of mehods, following
    Ref.~\cite{ogata_linear_1982}, builds upon the direct maximization of the
    likelihood function $\logl(\bm\theta | \bm N_t)$. This problem is concave, so
    efficient solvers like BFGS are able to quickly find a solution. Indeed, 
    the main drawback for this approach is the computational cost of each of the iterations
    of the algorithm, as the complexity of each evaluation of the gradient function
    scales with the number of points in the sample $n$.
    More precisely, this strategy requires to minimize the function
    \begin{equation}
        \label{eq:max_lik}
        -\logl(\bm\theta | \bm N_t) = \mathcal{H}(\bm\theta) = \sum_{i=1}^d {\bm h}^\top {\bm \theta} - \sum_{m=1}^n \log \sum_{a=0}^{dp} \theta^{u_m a} G^{ma} \, .
    \end{equation}
    where
    \begin{equation}
       G^{ma} = \int_0^T \dd t \, g^{a} (t_m - t)
    \end{equation}
    Even though one can reduce the cost of each iteration by precomputing the coefficients $\bm{\mathsf G}$ on the right-hand-side
    of Eq.~(\ref{eq:max_lik}) (complexity is $O(d^2 p \Lambda^2 T t$)), one is still left with a sum over all the $n$
    terms on the right hand side. This can have a huge impact on the
    computational limits of this strategy (see Fig.~\ref{fig:time_perf}), which quickly become prohibitive in $d$.
    In fact, the cost per evaluation of both the likelihood function and $\mathcal L(\bm N_t,\bm\theta)$ its gradient scales as $O(d^2 p \Lambda T)$. Then, if number of gradient evaluations is much larger than $dp$, then
    mean-field methods are faster. This is typically the case, as seen in Fig.~\ref{fig:time_perf}.

\subsection{Expectation-Maximization (EM)}
    \label{sec:EM}
    In the case of the model we focus on, EM is actually as complex as MLE, as it requires calculating
    a gradient of the likelihood function at each iteration~\cite{Lewis:2011aa}. According to the type of MLE algorithm adopted, the relative speed of 
    EM with respect to MLE can change. Taking BFGS as an example, the EM algorithm suffers actually from a slower convergence, as it is
    unable to exploit second-order information, as done instead by Newton and quasi-Newton methods do.

\subsection{Contrast Function (CF)}
    \label{sec:CF}
    The closest method to the one that we have presented is a generalization of the constrast function approach proposed
    in~\cite{reynaud2013inference}. In this framework, the $\bm\theta$ should minimize a loss function defined as
    \begin{equation}
        \mathcal{C}(\bm\theta) = \frac{1}{2} \sum_i \int_0^T \dd t \, \left( \frac{\dd N^i_t}{\dd t} - \lambda^i_t \right)^2 \, ,
    \end{equation}
    resulting in
    \begin{equation}
        \mathcal{C}(\bm\theta) = T \left(
    \sum_{i=1}^d ( -\bar\Lambda^i \bm k^{i\top} \bm\theta^i + \frac{1}{2} \bm\theta^{i\top} \bm{\mathsf{J'}} \bm\theta^i) \right) \, ,
    \end{equation}
    where the collection $\bm k^i$ is defined as in Eq.~(\ref{eq:aux_k}), while the matrix $\bm{\mathsf{J'}}$ is given by
    \begin{equation}
        {J'}^{ab} = \frac{1}{T} \int_0^T \dd t \, \dd N^a_{t'} \, \dd N^b_{t''} \, g^{a}(t-t') g^{b}(t-t'') \, .
    \end{equation}
    This approach also maps the inference problem on a linear system, whose solution is given by
    \begin{equation}
        \bm\theta^{i} = \bar\Lambda^i \bm{\mathsf{C'}} \bm{k}^{i}
    \end{equation}
    where $\bm{\mathsf{C'}} = (\bm{\mathsf{J'}})^{-1}$.
    This solution is numerically close to Eq.~(\ref{eq:estimator_theta}). The computation of the $\bm{\mathsf{J'}}$
    has a complexity of $O(d^2 p^2 \Lambda^2 T t)$, but due to the impossibility of exploiting the trick illustrated in App.~\ref{app:aux_func} for the computation of $\bm{\mathsf{J}}^i$ it needs to be implemented in a different manner (unless one doesn't introduce some artificial binning in the data). In practice (see Fig.~\ref{fig:time_perf}), we find MF to be significantly faster than CF. Nevertheless, CF seems to be a method particularly suitable in order to investigate the mean-field regime of the Hawkes process, although its precise relation with MF still needs to be rigorously established.

\section{Numerical examples}
\label{sec:simulation}
We report in this section the results obtained by calibrating a Hawkes process from synthetic data generated by a known model, and by comparing the performance of our algorithm with the one of known methods.

\paragraph{Single exponential basis kernel}
First, we have considered as a benchmark the case in which $p=1$, so that $g^{a} (t) = g_1 (t) \equiv g (t)$. For $j>0$, we have assumed the basis kernel to have the exponential structure
\begin{equation}
    \label{eq:exp_kern}
    g (t) = \beta e^{-\beta t} \ind{t>0} \, .
\end{equation}
The topology of the matrix $\bm\alpha$ has been chosen from the following ensemble:
\begin{itemize}
    \item A block structure in which each nodes belong to one out of several clusters. In particular we studied the case in which $\alpha^{ij}$ is zero (if $i$ and $j$ do not belong to the same cluster) or $\alpha^{ij} = \alpha / c $ (if they do belong to the same cluster, with size given by $c$);
\end{itemize}
Regardless of the structure, in both cases one has $||\bm \Phi ||_1 =||\bm\alpha||_1 = \alpha$, so that we used the parameter $\alpha\in \mathbb{R}_+$ in order to interpolate between the non-interacting case $\alpha=0$ and the critical one $\alpha = 1$. We have chosen for the purpose of these numerical experiments the values $\mu = [1,\dots, 1]$.

\begin{figure}[h]
    \begin{center}
        \includegraphics[scale=0.9]{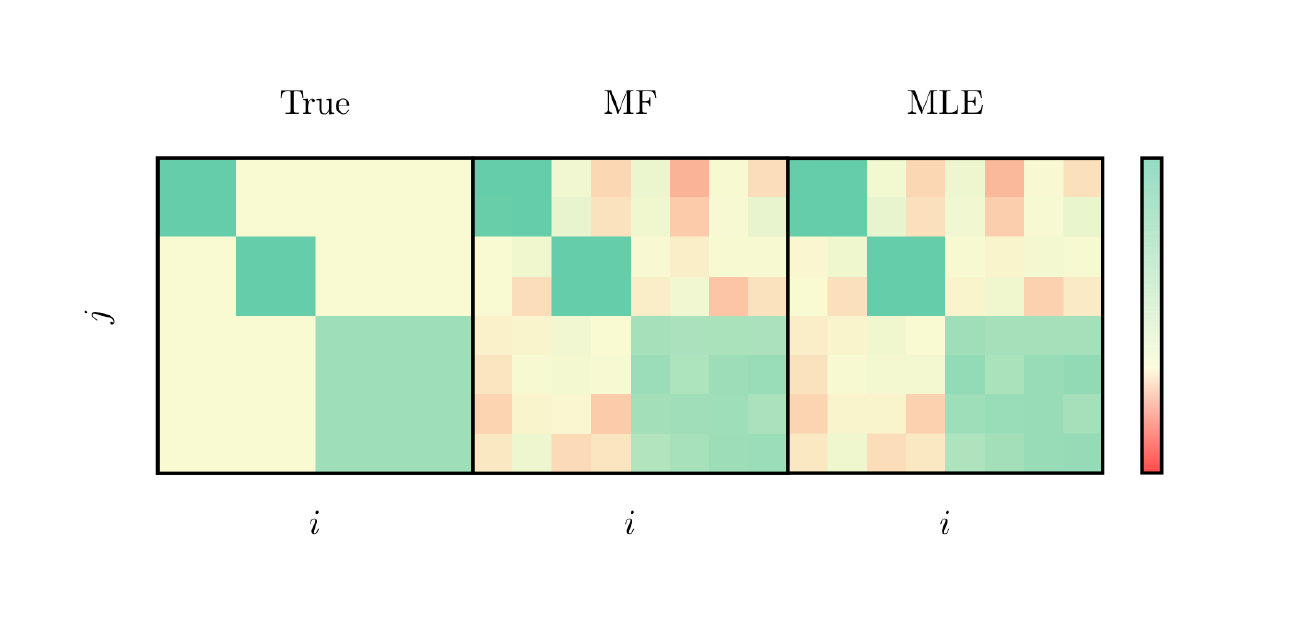}
    \end{center}
    \caption{Illustrative plot of the results of the reconstruction of
    a Hawkes model with a three-block structure for the matrix $\bm\alpha$.
    We used the parameters $||\bm\Phi||_1=0.5$, $\mu=1$ and $\beta = 1$, and
    ran the simulation for $T=10^4$. The true underlying matrix (left) is
    reconstructed both with the MF algorithm (center) and with a maximum
    likelihood estimation (right), yielding very similar results.}
    \label{fig:illustrative}
\end{figure}

As an illustrative example, we first plot in Fig.~\ref{fig:illustrative} the results of our algorithm in the case of a three block structure for the $\bm\alpha$ matrix. The results are for this choice of value essentially the same as the ones obtained by maximum likelihood estimation, with a considerable reduction in the computational time.

\begin{figure}[h]
    \begin{center}
        \includegraphics[scale=0.8]{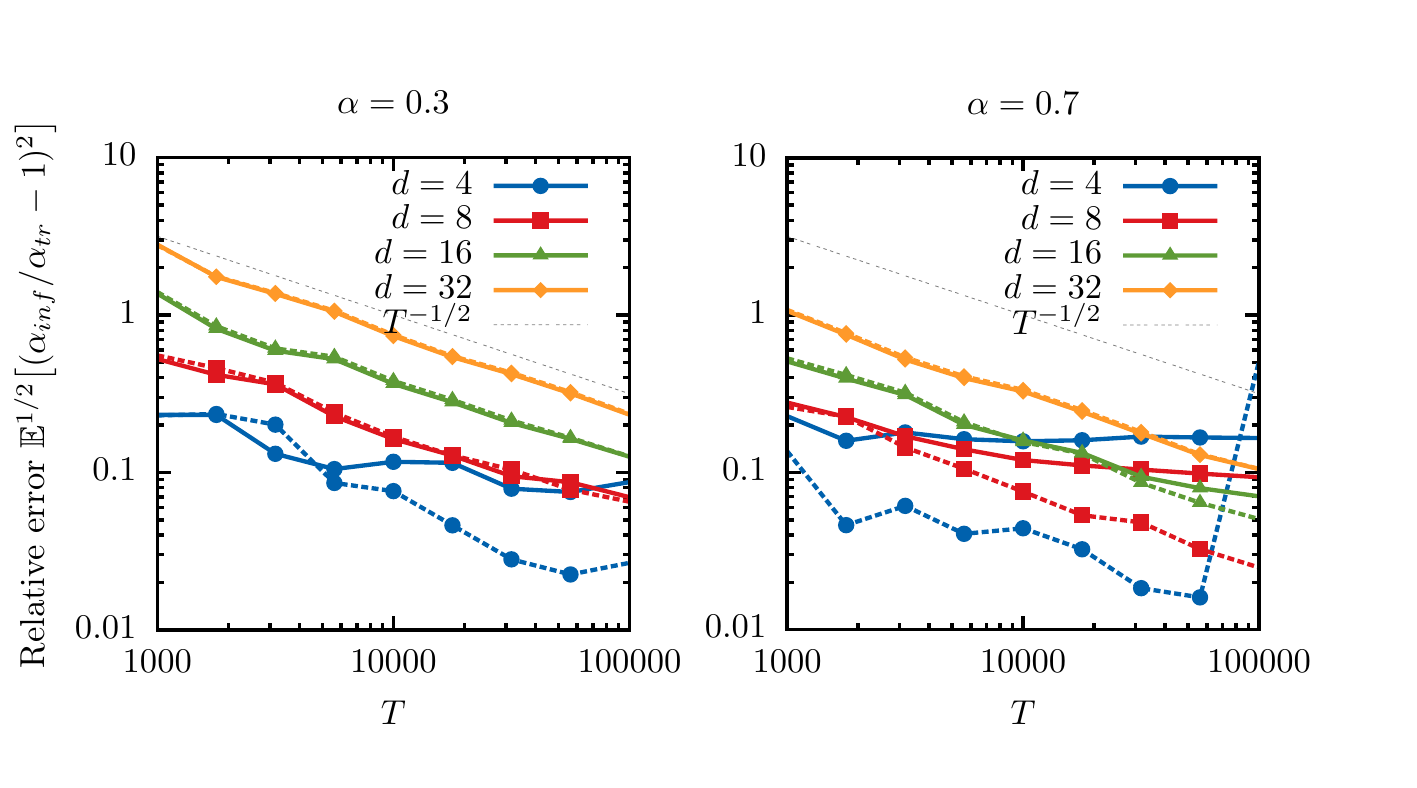}
    \end{center}
    \caption{Relative error on the non-zero couplings $\delta\alpha_{rel}$ as a function of the sample size $T$ for the maximum likelihood estimator (dashed lines) and the MF approximation (solid lines) for a fixed value of the empirical intensity $\bar{\bm\Lambda}$ in the two-block case as described in Sec.~\ref{sec:simulation} for various values of $\alpha$ and $d$. The other paramters are set to $\mu=\beta=1$.
    The figure shows that even  for moderate values of $d$ the error related to the MF
    approximation is much smaller than the statistical error in the estimation of the couplings
    induced by the finite value of $T$ affecting the maximum likelihood estimator. By comparing the left and the right panel, one sees that large values of $||\bm\Phi||_1=\alpha$ worsen the quality of the MF algorithm.}
    \label{fig:err_vs_T}
\end{figure}

\begin{figure}[h]
    \begin{center}
        \includegraphics[scale=0.8]{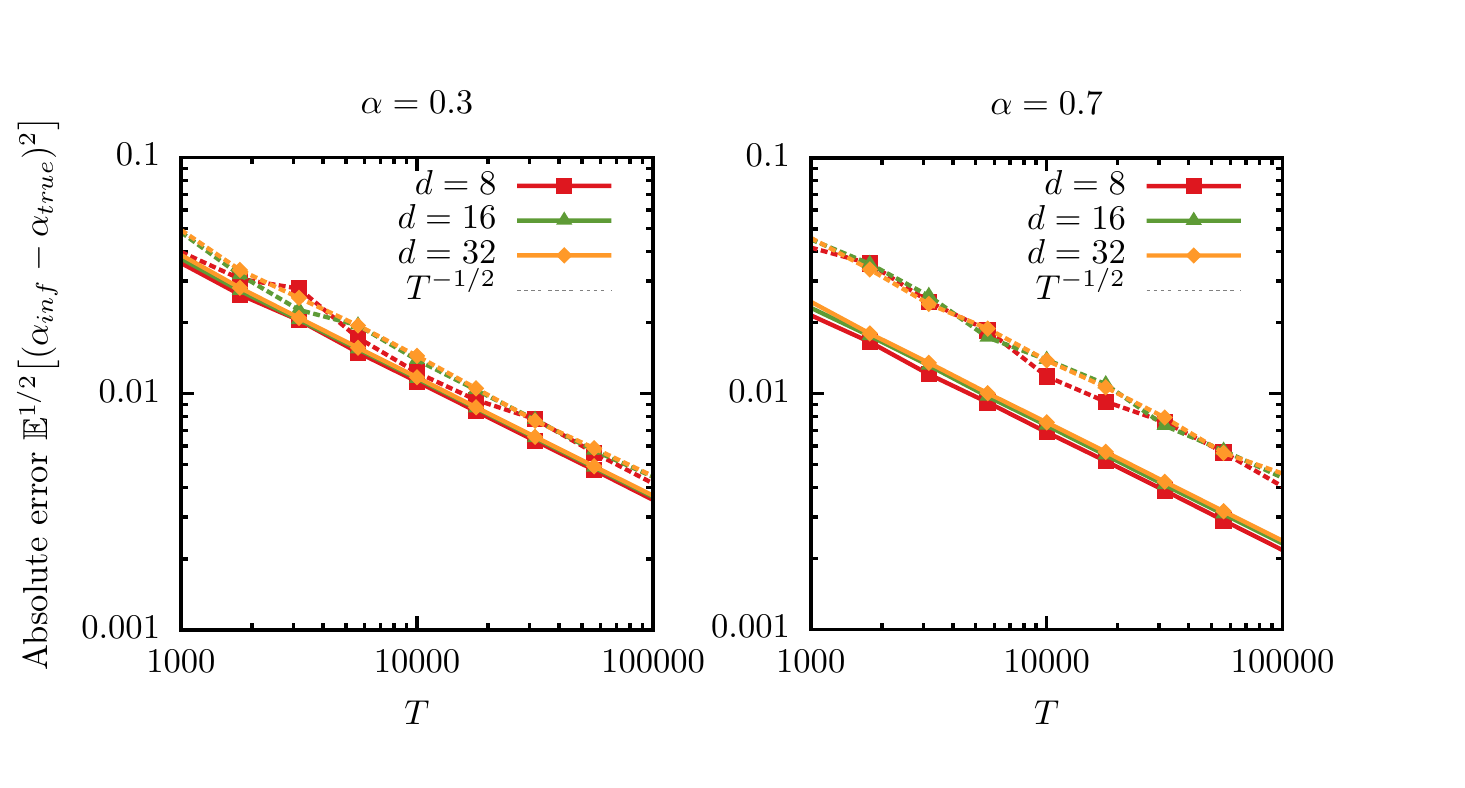}
    \end{center}
    \caption{Absolute error couplings $\delta\alpha_{abs}$ as a function of the sample size $T$ for the maximum likelihood estimator (dashed lines) and the estimation of the error provided by Eq.~(\ref{eq:covariance_theta}) (solid lines), for a fixed value of the empirical intensity $\bar{\bm\Lambda}$ and the same choice of parameters as in the previous figure.
    The figure shows that the mean-field estimation of the absolute error is close to the one find by maximum-likelihood, and that the mean-field estimator of the covariance underestimates the error, that grows when approaching the critical point.}
    \label{fig:error_estimation}
\end{figure}

In order to systematically assess the performance of the algorithm under various
circumstances, we have varied $\alpha$ uniformly in the interval $[0,1)$,
and addressed the problem of studying the scaling of the results in $d$
(we used $d = 2,4,8,\dots, 128$) and in $T$ (we considered $T\in [10^3,\dots,10^5]$).
Unless specifed otherwise throughout
the text, we have assumed block structures with two clusters, so that $c \in
\{
    \left\lceil{d/2}\right\rceil,\left\lfloor{d/2}\right\rfloor
\}$.
We measured the quality of the reconstruction by using
\begin{itemize}
    \item The negative log-likelihood of the sample
    $\mathcal{L}(\bm N_t,\bm\theta)= -\log\prob(\bm N_t|{\bm\theta}_{inf})$
    under the inferred couplings
    \item The relative error on the non-zero couplings
    \begin{equation}
        \label{eq:rel_err}
        \delta\alpha^2_{rel} = \sum_{i,j | \alpha^{ij}_{true} \neq 0} \left( \frac{\alpha^{ij}_{inf}}{\alpha^{ij}_{true}} - 1\right)^2 \, .
    \end{equation}
    \item The absolute error on all the couplings
    \begin{equation}
        \label{eq:abs_err}
        \delta\alpha^2_{abs} = \sum_{i,j} \left( \alpha^{ij}_{inf} - \alpha^{ij}_{true}\right)^2 \, .
    \end{equation}
\end{itemize}
The results that we have found for the relative error $\delta\alpha_{rel}$ in the two-block
case are summarized in Fig.~\ref{fig:err_vs_T}, where we have shown them for $\alpha = 0.3$
and $\alpha = 0.7$. What one finds is that, as expected, the error of the maximum
likelihood estimator decreases as $T^{-1/2}$ by increasing the sample size $T$. Indeed, the same
plot also shows that the MF estimator is able to match the results of the maximum likelihood
estimator up to a maximum value $T_\star$, beyond which the statistical error no longer dominates
$\delta\alpha_{rel}$ for the mean-field estimator, and the performance of MF deteriorates.
At that point, the approximation error becomes comparable with the statistical one, inducing a plateau in the curve of the error as a function of $T$.
What is interesting is that --~even for moderate values of $d$~-- the value of $T_\star$ is quite
large, indicating that the MF approximation is extremely effective in a broad range of regimes.
By increasing the value of $||\bm \Phi||_1=\alpha$, even though the relative error decreases due to the increase of the signal $\alpha$, the quality of the MF approximation decreases, and the value of $T_\star$ becomes smaller. Indeed, for a fixed value of $T$ and $||\bm\Phi||_1$, it is always possible to find a sufficiently large $d$ so that $T \ll T_\star$, and the MF approximation is effective.
This is confirmed in Fig.~\ref{fig:error_estimation}, where we have represented the absolute error $\delta\alpha_{abs}$ obtained by MLE against the one estimated by using Eq.~\ref{eq:covariance_theta}.

\begin{figure}[h]
	\begin{center}
		\includegraphics[scale=0.8]{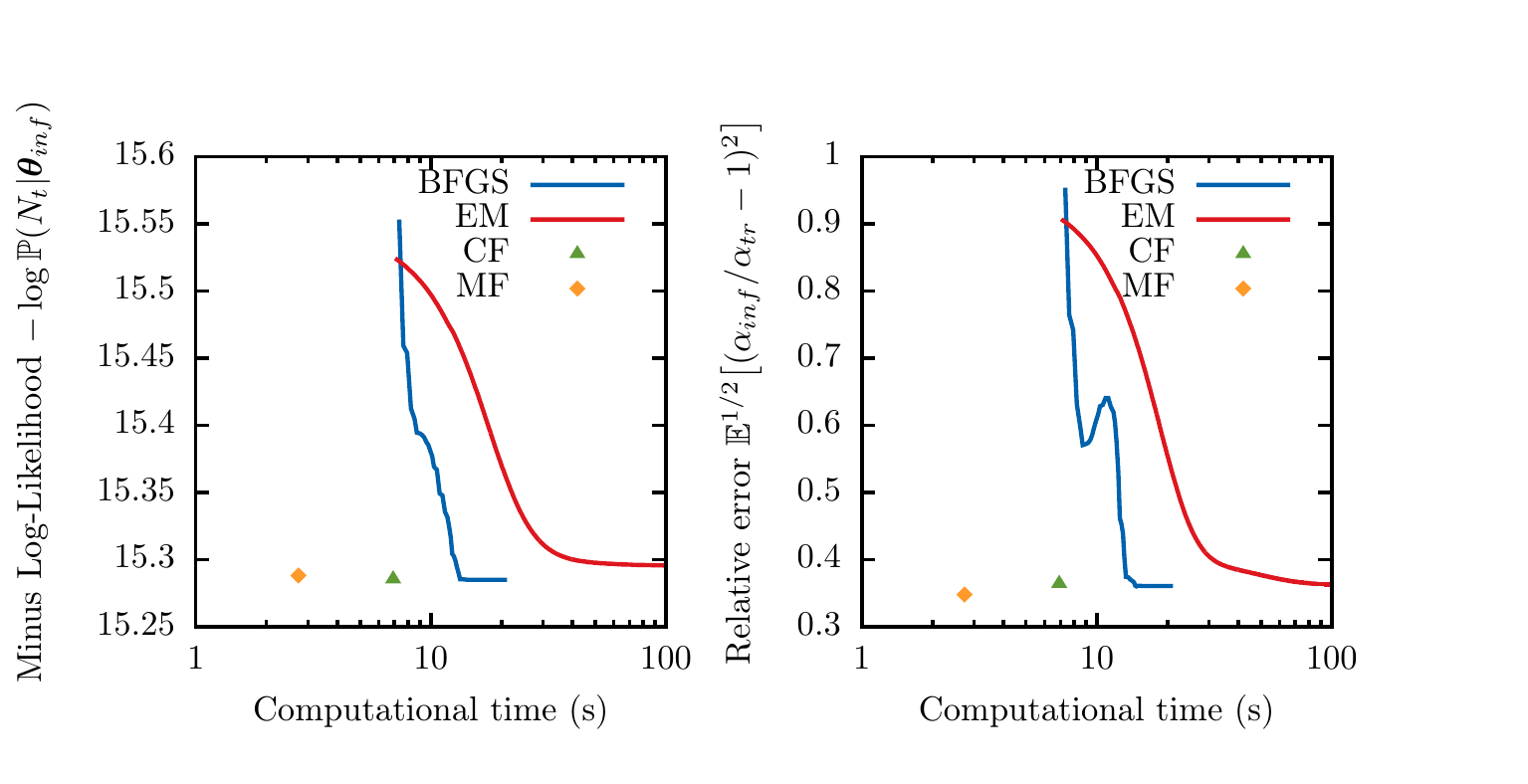}
	\end{center}
	\caption{Time required by the learning algorithms described in Sec.~\ref{sec:complexity} in order to achieve fixed values of accuracy, as measured by the negative log-likelihood (left panel) and the relative error (right panel).
		We have considered the two block case described in Sec.~\ref{sec:simulation}, and adopted the same parameters as in Fig.~\ref{fig:err_vs_T}.}
	\label{fig:time_perf}
\end{figure}

On the other hand, the computational time required in order to run the MF algorithm is considerably
smaller, as summarized in Fig.~\ref{fig:time_perf}.
In such figure we have represented the computational time required in order to run the different algorithm up to a target negative log-likelihood $ \mathcal{L}(\bm{N}_t,\bm\theta)$, on a machine with the same specifications as above, and shown that by taking into account both the pre-processing phase and the
solution of the linear system via matrix inversion, the MF algorithm achieves within a single iteration a value of the target function which is very close to the
the asymptotic one got from a BFGS minimization.

\begin{figure}[h]
	\begin{center}
		\includegraphics[scale=0.8]{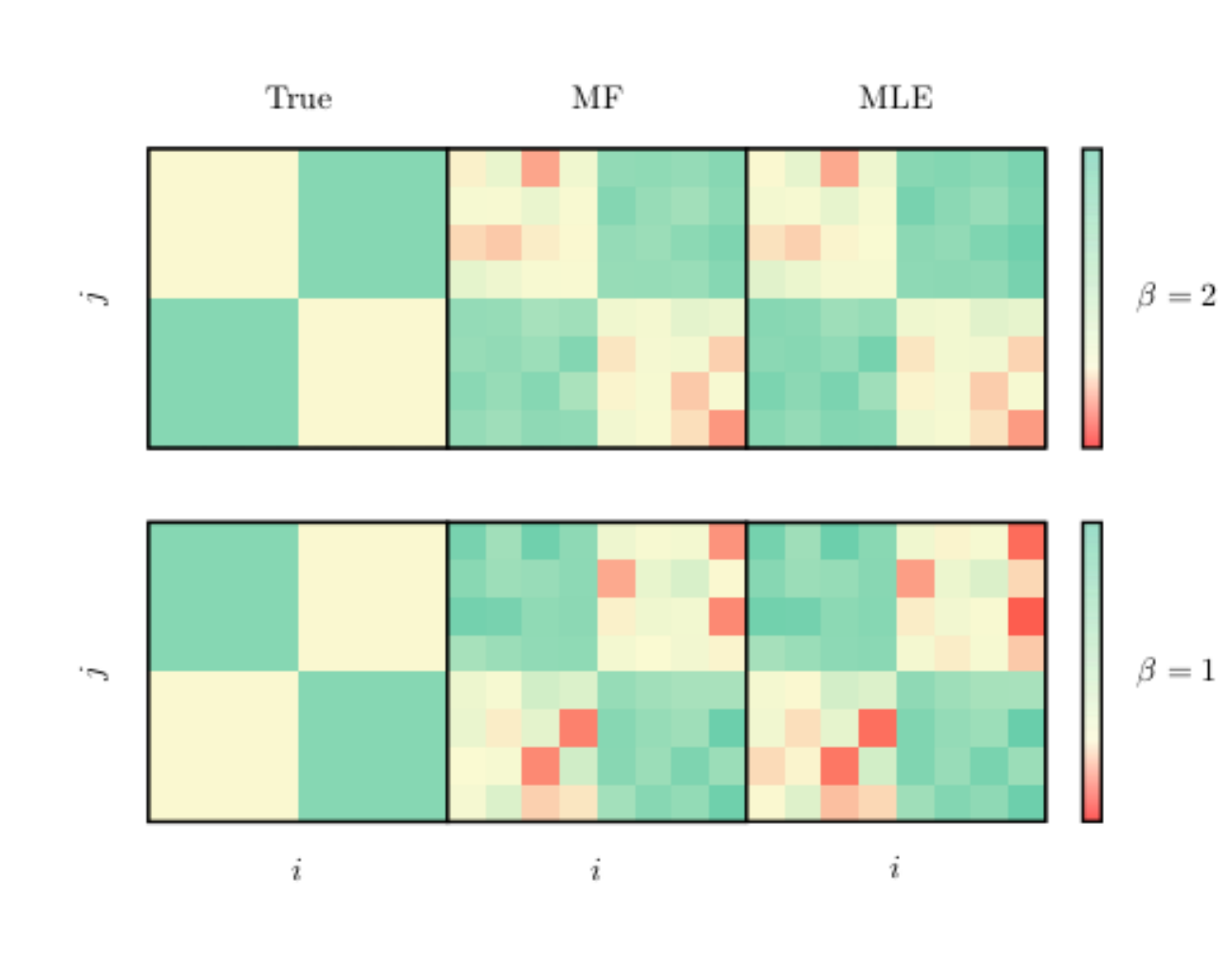}
	\end{center}
	\caption{Reconstruction of a Hawkes process in which two different exponential kernels
		are mixed together. In this example we have chosen $d=8,p=2$, and we have chosen the matrices
		${\bm\alpha}_1$ and ${\bm\alpha}_2$ as in the left panels of the figure. They both
		have a block structure with uniform entries equal to either zero or $\alpha = 0.5/c$,
		where $c=4$ is the connectivity of the block. The other parameters are set to $\mu=1$, $\beta_1 = 1$,
		$\beta_2 = 2$ and $T=10^5$. Also in this case our MF algorithm (central panels)
		is able to reconstruct  the interaction network approximately as well as the
		maximum likelihood estimators (right panels).}
	\label{fig:multi_exp}
\end{figure}

\begin{figure}[h]
	\begin{center}
		\includegraphics[scale=0.8]{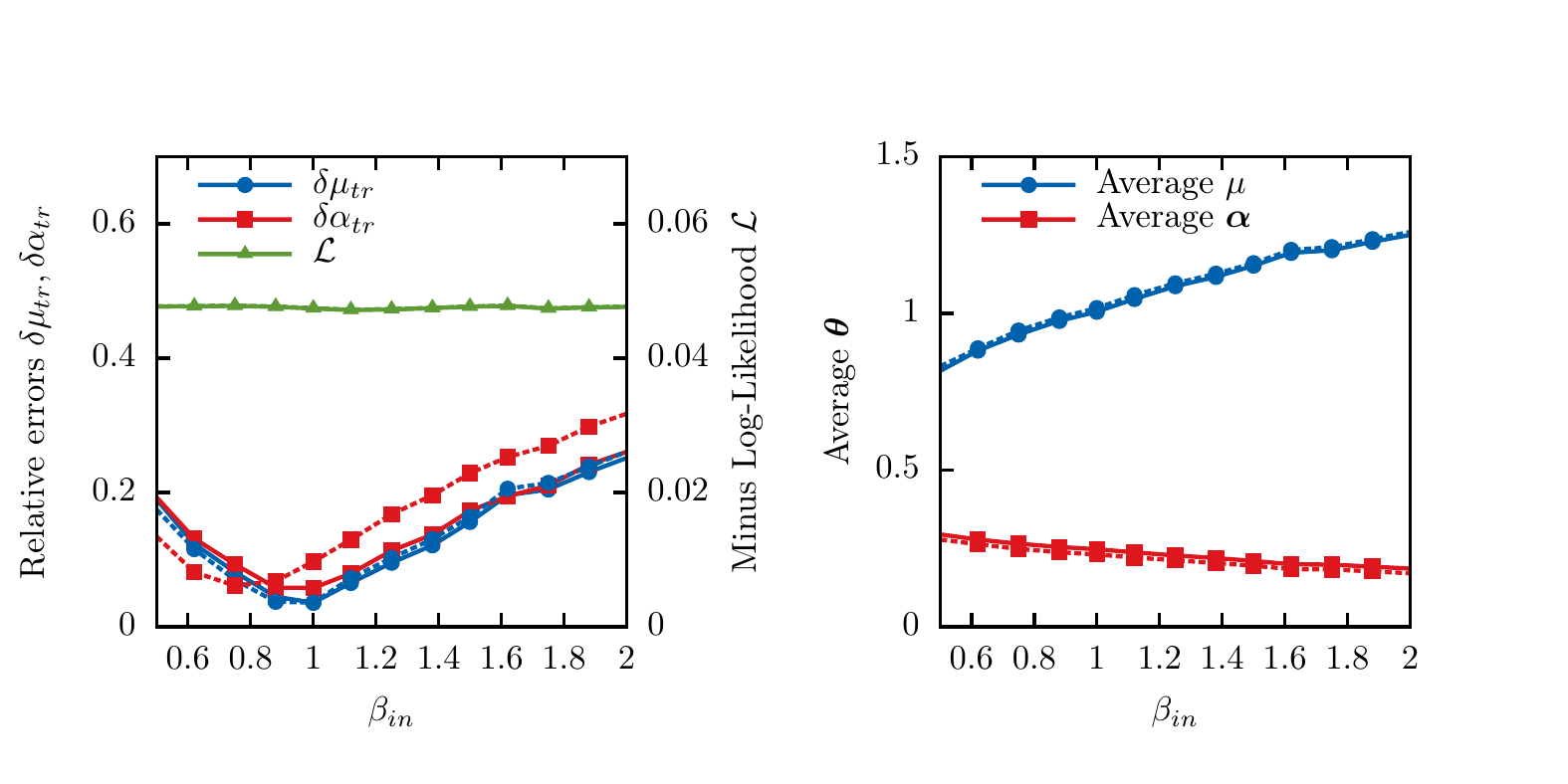}
	\end{center}
	\caption{Effect of the mispecification of $\beta$ on the quality of inference. We simulated
		a two-block structure for the matrix $\bm\alpha$ as described in Sec.~\ref{sec:simulation}, with $d=8$, $\mu=\beta_{true} =1$ and $T=10^5$, where $\beta_{true}$ denotes the true value of the
		decay constant $\beta$. We subsequently inferred the parameters $\mu$
		and $\bm\alpha$ under the maximum likelihood estimator (solid line) and the MF one
		(dashed line) under various values for the decay constant $\beta_{in}$ of the exponential kernel $\bm g_{in}(t)$. We find that, even though the likelihood is barely changed by the shift in
		$\beta_{in}$ (left panel), its change reflects in a bias in the inferred couplings (right panel).}
	\label{fig:beta_dep}
\end{figure}

\paragraph{Multiple exponential basis kernels}
We have also considered the more general case in which the $\bm{\Phi}(t)$ is
given by a sum of kernels of different nature. Even though the implementation of
an algorithm with multiple basis kernels is technically more challenging, there
are no qualitative differences with respect to what we have stated above.
That is why we have chosen to present an illustrative example showing how the
MF algorithm performs in the multiple basis case.
In particular, we have considered an example in which $p=2$, in which the matrix
${\bm\alpha}_1$ has a uniform two-block structure as in the previous case, while
for the second matrix ${\bm\alpha}_2$ only the complementary blocks are uniformly
filled (see Fig.~\ref{fig:multi_exp}). For both ${\bm\alpha}_1$ and ${\bm\alpha}_2$,
the non-zero entries have value $\alpha / c$, being $c$ the size of the block.
Finally, we chose $\beta_1^{ij} = 1$ and $\beta_2^{ij} > 1$, so that for
${\bm\beta}_1={\bm\beta}_2$ one recovers a completely homogeneous interaction
structure.
Fig.~\ref{fig:multi_exp} shows an illustrative plot of the results of the
reconstruction in the multiple basis kernel scenario.

\paragraph{The role of $\beta$}
We have also studied the effect of the misspecification of $\beta$ by studying the 
effect of an input kernel $\bm{g}_{in}(t)$ different from the $\bm{g}_{true}(t)$ used
to generate the data. We did it in the case of $p=1$, with a block $\bm\alpha$ matrix
as described in the paragraph above relating to the single exponential basis kernel scenario.
Also in this case we have chosen $\mu=1=\beta_{true}=1$, with $\beta_{true}$ denoting the true and unique decay constant for the exponential kernels, whereas $\beta_{in}$ will denote the decay
constant chosen to perform the inference.
Our results are summarized in Fig.~\ref{fig:beta_dep}, where we illustrate them in a specific case.
We find that the likelihood of the model under the inferred couplings is not strongly affected
by the misspecification of $\beta$, as the inferred intensity turns out to be very weakly dependent
on $\beta_{in}$. What actually happens is that by changing $\beta_{in}$ is is possible to tune
a bias changing $\bm\mu$ in favor of $\bm\alpha$.

This behavior is easy to understand in the framework of our
MF approximation. In fact, App.~\ref{app:small_fluct} shows that the value of the inferred couplings $\bm\alpha$ is proportional to the fluctuations of the auxiliary function $\bm{\delta k}$. According to Eq.~\eqref{eq:k_cum_exp}, these fluctuations are obtained by integrating the empirical correlation function against against the kernel $\bm g_{in}(t)$ in order to obtain $\bm{\delta k}$. Hence, when $\beta_{in}$ is large $\bm g_{in}(t)$ is peaked close to the origin, where the correlations are larger, resulting in a greater value for $\bm{\delta k}$. Conversely, for large $\beta_{in}$ such integral assumes a small value, because the correlations are small at the timescales in which $\bm g_{in}(t)$ has large mass.
Hence, small value of $\beta_{in}$ leads to a bias in favor of $\mu$, while large values favor $\bm\alpha$. Independently of $\beta_{in}$, the combination of $\mu$ and $\bm\alpha$
leading to the inferred value of $\bm \Lambda$ is almost unchanged.

Even though these conclusions depend on the choice of an exponential kernel for
$\bm g_{in}$, in more general cases it is possible to predict the direction of the bias
by checking the timescales in which the mass of the correlation function lies agains the
timescales relative to $\bm g_{in}$. A maximum overlap favors $\bm\alpha$, while a small one
will increase $\mu$.


\section{Conclusion and prospects}
\label{sec:conclusions}
We have presented an approximated method for the 
calibration of a Hawkes process from empirical data. This method allows one to obtain in close form
the inferred parameters for such a process and its implementation is considerably faster than
the algorithms customarily employed in this field (MLE+BFGS, EM, CF). In particular, we map the problem of the estimation of a Hawkes process on a least-square regression, for which extremely efficients techniques are
available~\cite{dennis1977quasi,liu1989limited}. Although this method is asymptotically biased,
we have shown numerically and analytically that the bias is negligible
in situations when the interactions of the system are sufficiently weak or self-averaging, which
is thought to be the case in many applications to high-dimensional inference.

We believe that our method is not limited to the framework that we have presented, but can be extended to more general settings like marked or non-linear Hawkes processes.
Another prospect of  particular interest is the case in which a non trivial prior is added to the Bayesian
formula Eq.~(\ref{eq:bayes}), which allows one to embed {\bf regularization} in our framework.
Popular priors that are relevant in the field of optimization include for example the Gaussian prior
\begin{equation}
\prob_0^{L^2}(\bm \theta) \propto \exp\left( \sum_{ia} \kappa^{ia} (\theta^{ia})^2 \right)
\end{equation}
and the Laplace prior
\begin{equation}
\prob_0^{L^1}(\bm \theta) \propto \exp\left( \sum_{ia} \kappa^{ia} |\theta^{ia}| \right) \, .
\end{equation}
The inference problem analogous to the evaluation of the partition function~(\ref{eq:part_func}) associated with
these priors is obviously more complex than the one with a flat prior that we have presented.
Nevertheless, our mean-field approach allows even in those cases a mapping on simpler problems,
for which efficient solution methods exists~\cite{liu1989limited}. In particular,
\begin{itemize}
	\item The case of a Gaussian prior can be mapped to the one of the minimization of the quadratic form
	\begin{equation}
	\mathcal{H}^{L^2}(\bm\theta) 
	= \sum_{i=1}^d \left( \frac{1}{2}\bm\theta^{i\top} \bm{\mathsf{J}}^i\bm\theta^i - (2\bm{k}^i - \bm h)^\top \bm\theta^i \right)
	+ \sum_{ia} \kappa^{ia} (\theta^{ia})^2
	\end{equation}
	\item The case of a Laplace prior can be mapped to the LASSO problem~\cite{tibshirani1996regression}
	\begin{equation}
	\mathcal{H}^{L^1}(\bm\theta)
	= \sum_{i=1}^d \left( \frac{1}{2}\bm\theta^{i\top} \bm{\mathsf{J}}^i\bm\theta^i - (2\bm{k}^i - \bm h)^\top \bm\theta^i \right)
	+ \sum_{ia} \kappa^{ia} |\theta^{ia}|
	\end{equation}
\end{itemize}
Other types of regularizers might be considered by applying this same scheme.

We believe the approach we propose in this paper to be of particular interest for the problem of high-dimensional inference as, by construction, it is particularly effective in the regime of large $d$ (especially relevant for big-data analysis) where traditional numerical methods can be computationally expensive.
Last but not least, the mean-field method can provide an easy and efficient way to set a good starting point 
for standard iterative algorithms that maximize the likelihood objective.

\section*{Acknowledgements}
This research benefited from the support of the ``Chair Markets in Transition'',
under the aegis of ``Louis Bachelier Finance and Sustainable Growth'' laboratory,
a joint initiative of \'Ecole Polytechnique, Universit\'e d'\'Evry Val d'Essonne
and F\'ed\'eration Bancaire Fran\c{c}aise.


\appendix

\section{Fluctuation ratio of the intensity on the homogeneous case} 
\label{sec:homogeneous_system}
In order to get an intuition about the behavior of a Hawkes process in a very stylized setting, we present an estimation of the square root of the ratio of the variance of ${\bm \lambda_t}$ over its squared mean. Thus, we will localize in the phase space of the model the points in which the regime of large dimension $d$ (quasi-homogeneous) and/or weak interactions (see (i) and (ii) in Sec. \ref{sec:3.3}) can be detected. In order to do this, we study the quantity 
\begin{equation}
    r_i = \sqrt{\frac{\var(\lambda_t^i)}{\expect^2(\lambda^i_t)}} \, ,
\end{equation}
in the simplest possible framework, that is, a perfectly homogeneous Hawkes process of the form
\begin{eqnarray}
    {\bm \mu} &=& \mu \, [1,\dots, 1] \\
    \bm{\Phi}(t) & = & \Phi(t) \, [1]_{1\leq i,j \leq d}
\end{eqnarray}
Let us point out that in this appendix we calculate the observables of a Hawkes process by averaging over its stationary measure, instead of using the empirical averages. Hence, when using these result in Sec.~\ref{sec:3.3}, one should keep in mind that we are replacing all the empirical averages by expectations. 

\vskip .2cm 
\paragraph{1. Mean of ${\bm \lambda_t}$} 
The mean value is easy to compute once one particularizes the results of \cite{Hawkes:1971lc,Hawkes:1971nq} in the homogeneous case, obtaining
\begin{eqnarray}
    {\bm \Lambda}
    &=& \expect {\bm \lambda_t} = \left(\id - \left[\int \Phi^{ij}(t)\dd t\right]_{1\le i,j \le d}\right)^{-1} {\bm \mu} \\
    &= &\left(\id - ||\Phi||_1  [1]_{1\le i,j \le d} \right)^{-1} {\bm \mu} \\
        &= &\frac {\mu}{1-d||\Phi||_1} [1]_{1\le i\le d}, \\
\end{eqnarray}
Let us point out that we want to consider the focus on a regime in which the stability condition {\bf (SC)} hold,
so that we hold the averaged intensity ${\bm \Lambda}$ constant when varying the dimension $d$.
In order to do so, we choose to keep the exogeneous intensity $\bm\mu$ constant and let $\Phi(t)$ vary so that $d ||\Phi||_1$ is constant.
Thus we set
$$ \Phi(t) = g(t) ||\bm \Phi||_1/d,~~~\mbox{with}~~||g||_1 = 1.$$ 
In that case, one gets
\begin{equation}
    {\bm \Lambda} = \left[ \frac{\mu}{1-||\bm\Phi||_1} \right]_{1\leq i \leq d}= [\Lambda]_{1\leq i \leq d}\,
\end{equation}

\vskip .2cm
\paragraph{2. Variance of ${\bm \lambda_t}$}
The variance can be obtained by using the stationary version of the martingale representation for the Hawkes process (see~\cite{bacry2012non})
\begin{equation}
    {\bm \lambda_t} = {\bm \Lambda} + \int_{-\infty}^t \bm\Psi(t-t') \, \dd {\bm M_{t'}} \, ,
\end{equation}
where ${\bm M_t}$ is the martingale ${\bm M_t} = {\bm N_t} - \int_0^t {\bm \lambda_t} \dd t$, (one has $\expect[\dd {\bm M_t} | \mathcal{F}_t] = 0$), and $\bm\Psi(t)$ can be obtained from the Fourier transform of $\bm{\Phi}(t)$, that we denote as $\hat{\bm\Phi}(\omega) = \int \dd t {\bm\Phi}(t)e^{-i\omega t}$:
\begin{equation}
    \hat{\bm\Psi}(\omega) =  (\id - \hat{\bm\Phi}(\omega))^{-1} - \id ,
\end{equation}
as shown in~\cite{Hawkes:1971lc,Hawkes:1971nq} and generalized in~\cite{bacry2012non}. Then one can write
\begin{eqnarray}
    \var(\lambda_t^i) &=& \sum_{1\leq j,k \leq d} \expect\left[ \int_{-\infty}^t \dd M_s^j \, \int_{-\infty}^t \dd M_{s'}^{k} \Psi^{ij}(t-s) \Psi^{ik}(t-s')\right]  \\
& = & \sum_{j=1}^d \Lambda^j \int_0^\infty  \dd t \, (\Psi^{ij})^2(t)
= d \Lambda ||\Psi||^2 = \frac{d \Lambda}{2\pi} ||\hat \Psi||^2 ,
\end{eqnarray}
where $||.||$ stands for the $L^2$ norm. However, since
$$
    \hat{\bm\Psi}(\omega) 
    =  (\id - \hat{g}(\omega)||\bm \Phi||_1[1]_{1\leq i,j \leq d})^{-1} - \id \\
    = \frac {\hat{g}(\omega)||\bm \Phi||_1}{1-\hat{g}(\omega)||\bm \Phi||_1}   [1]_{1\leq i,j \leq d},\\
$$
one has
$$
\hat \Psi(\omega) 
= \frac {\hat{g}(\omega)||\bm \Phi||_1}{1-\hat{g}(\omega)||\bm \Phi||_1}.
$$
Thus
\begin{equation}
\label{eq:vr}
        \var(\lambda_t^i)  = 
    \frac{\Lambda ||\bm \Phi||_1^2} d \int \dd \omega 
\frac{|\hat{g}(\omega)|^2}{\left|1-||\bm \Phi||_1 \hat{g}(\omega)\right|^{2}}. 
\end{equation}
So the scaling in either limit $||\bm \Phi||_1 \rightarrow 0$
(case (i) in Sec. \ref{sec:3.3}) or $d\rightarrow +\infty$ (case (ii) in Sec. \ref{sec:3.3})
is thus of the form $\var(\lambda_t^i) \sim {\Lambda ||\bm \Phi||_1^2}/d$. Let's try to be more accurate and characterize the scaling when approching the critical case  $||\bm \Phi||_1 \rightarrow 1$.
It will naturally depend on the behavior of $\hat g(\omega)$ around 0.
We set
$$
\hat g(\omega) = 1+K\omega^{\eta}+ o(\omega^\eta).
$$
Let us point out that in the case $g$ is exponential, one has $\eta = 1$ and in the case $g(x) \sim x^{-\gamma}$ (when $x \rightarrow +\infty$) 
then, using a Tauberian theorem, one can prove that $\eta = \gamma+1$. 
Let's set $\epsilon = 1-||\bm \Phi||_1$. Then, focusing on $|\omega|<a$ ($a$ fixed), the integral in \eqref{eq:vr} becomes
\begin{eqnarray}
\nonumber
\int_{|\omega|<a} \dd \omega 
\frac{|\hat{g}(\omega)|^2}{\left|1-(1-\epsilon) \hat{g}(\omega)\right|^{2}} 
& \simeq & \int_{|\omega|<a} \dd \omega 
\frac{1}{\left|1-(1-\epsilon) (1+K\omega^\eta)\right|^{2}} \\
\nonumber
& \simeq &
 \int_{|\omega|<a} \dd \omega \frac{1}{\epsilon^2 + \Re{(K)}^2\omega^{2\eta} -2\Re{(K)}\epsilon\omega^\eta + \Im{(K)}^2\omega^{2\eta}} \\
\nonumber
& \simeq & 
\epsilon^{1/\eta-2}  \int_{|\omega|<a\epsilon^{-1/\eta}} \dd \omega \frac{1}{1 + \Re{(K)}^2\omega^{2\eta} -2\Re{(K)}\omega^\eta + \Im{(K)}^2\omega^{2\eta}} \\
\nonumber
& \sim &  \epsilon^{1/\eta-2}
\end{eqnarray}
This last result along with \eqref{eq:vr} gives that
\begin{equation}
        \var(\lambda_t^i)  = \frac{\Lambda}{\tau_g(||\bm \Phi||_1)}
    \left( \frac{\Lambda ||\bm \Phi||_1^2} {d (1-||\bm \Phi||_1)^{2-1/\eta}} \right) , 
\end{equation}
where 
$$
\tau_g(x) = (1-x)^{-2+1/\eta} \left( \int d\omega \frac {|\hat{g}(\omega)|^2}{\left|1-x \hat{g}(\omega)\right|^2} \right)^{-1}
$$ is a function on $[0,1]$ which is both bounded and bounded away from zero.
As it is illustrated below in the exponential case, the constant  $\tau_g(||\bm \Phi||_1)$ can be interpreted as a characteristic time scale of the kernel.

\vskip .2cm
\paragraph{3. Fluctuations of ${\bm \lambda_t}$ and {\bf (MFH)}}
Thus, the fluctuations of ${\bm \lambda_t}$ behave, in either limit (i) weak interactions ($||\bm \Phi||_1\ll1$, case (i) in Sec. \ref{sec:3.3}) or (ii) large dimension ($d\gg1$
 case (ii) in Sec. \ref{sec:3.3}), like 
\begin{equation}
\label{eq:fluc}
    \frac{\var(\lambda_t^i)}{\expect^2(\lambda^i_t)} \sim \frac{1}{\Lambda \tau_g}\left(\frac{||\bm \Phi||_1^2} 
    {d (1-||\bm \Phi||_1)^{2-1/\eta}}\right), 
\end{equation}
and the mean-field hypothesis simply writes
\begin{equation}
{\bf (MFH)~: }~~~~    r_i = \sqrt{\frac{\var(\lambda_t^i)}{\expect^2(\lambda^i_t)}}  \sim \frac{||\bm \Phi||_1} {\sqrt{d \Lambda \tau_g} (1-||\bm \Phi||_1)^{1-1/2\eta}} \ll 1,
\end{equation}

\vskip .2cm
\paragraph{4. Exponential case}
In the simpler case in which
\begin{equation}
    g(t) =  \beta e^{-\beta t} \ind{t>0} \,
\end{equation}
we get 
\begin{equation}
    \hat g(\omega) = \frac{1}{1+ i\omega/\beta}~~\mbox{and}~~\eta = 1 ,
\end{equation}
which implies
\begin{equation}
    \tau_g(||\bm\Phi||_1) = \frac 2 \beta \, .
\end{equation}
We see here that $\tau_g$ does correspond to the characteristic time scale of the kernel.
The mean-field hypothesis simply writes
\begin{equation}
\label{eq:flucexp}
 {\bf (MFH)~: }~~~~    r_i = \sqrt{\frac{\var(\lambda_t^i)}{\expect^2(\lambda^i_t)}}  \sim \frac{||\bm\Phi||_1\sqrt{\beta}}{\sqrt{2d\Lambda(1-||\bm\Phi||_1)}} \ll 1 \,.
\end{equation}
This result is illustrated in Fig.~\ref{fig:scaling_fluct}.


\section{Auxiliary functions}
\label{app:aux_func}
This appendix is devoted to a more detailed analysis of the auxiliary functions, so to provide a recipe for an efficient numerical implementation. We start by reminding that
\begin{eqnarray}
h^{a}  & = & \frac{1}{T}\int_0^T \dd t \int_{0}^t \dd N_{t'}^a g^{a}(t-t') \label{eq:app_h}\\
k^{ia}  & = & \frac{1}{N^i_T} \int_{t\neq t'} \dd N^i_t \dd N^a_{t'} g^{a}(t-t') \label{eq:app_k}\\
J^{iab} & = & \frac{T}{(N^i_T)^2} \int_{t\neq t',t''} \dd N^i_t \dd N^a_{t'} \dd N^b_{t''} \label{eq:app_j}\\ 
& \times & g^{a}(t-t') g^{b}(t-t'')
\end{eqnarray}
required in order to compute the quantities Eq.~(\ref{eq:estimator_theta})
and~(\ref{eq:covariance_theta}) defined in the main text.
First, note that while in order to compute Eq.~(\ref{eq:app_k}) one needs to discard the point $t=t'$, in the case of Eq.~(\ref{eq:app_j}) it is necessary to consider the points $t\neq t'= t''$, so that it reads more explicitly
\begin{eqnarray}
    J^{iab} & = & \frac{T}{(N^i_T)^2} \int_{t\neq t'\neq t''} \dd N^i_t \dd N^a_{t'} \dd N^b_{t''} \\ 
& \times & g^{a}(t-t') g^{b}(t-t'') \nonumber \\
& + & \frac{T}{(N^i_T)^2}
    \delta^{i_a i_b} \int_{t\neq t'} \dd N^i_t \dd N^a_{t'} g^{a}(t-t') g^{b}(t-t') \nonumber \, .
    \label{eq:app_alt_j}
\end{eqnarray}

\paragraph{Implementation} The non-zero components of the auxiliary functions can be computed efficiently by exploiting the relations
\begin{eqnarray}
    h^{ij}  & = & \frac{1}{T} \sum_{m=1}^n w^{ij}_{t_m} \\
    k^{ij}  & = & \frac{1}{N^i_T} \sum_{m=1}^n y^{ij}_{t_m} \\
    J^{ijk} & = & \frac{T}{(N^i_T)^2} \sum_{m=1}^n y^{ij}_{t_m} y^{ik}_{t_m}
\end{eqnarray}
where one has defined the quantities $w^{ia}_{t_m}$ and $y^{ia}_{t_m}$ as
\begin{eqnarray}
    w^{ij}_{t_m} & = & \int_{t_m}^T \dd t \, g^{ij}(t-t_m) \\
    y^{ij}_{t_m} & = & \sum_{m'=1}^{m-1} g^{ij}(t_m - t_{m'}) \, \delta^{i u_m} \delta^{j u_{m'}} \, ,
\end{eqnarray}
while for $a=0$ or $b=0$, one can use the relations $J^{i0a}=J^{ia0}=k^{ia}$ and $k^{i0} = \Lambda^i$.
Note that
\begin{itemize}
    \item The values of $w^{ij}_{t_m}$ and $y^{ij}_{t_m}$ don't have to be stored in memory,
    and can rather be computed on-line during the evaluation of the auxiliary functions.
    \item The basis kernels ${\bm g}(t_m-t_n)$ can be cut-off when they falls below a target value of $\epsilon$ in order to gain computational speed. As specified in Sec.~\ref{sec:inference} This implies that the computation of the auxiliary functions is of the order of
    $O(d^2 p \Lambda T \times \max(\Lambda t, d p))$, where $t$ is an $\epsilon$-depedent threshold. The $\max$ operation arises due to the computation
    of $y^{ij}_{t_m}$: while the computation of this intermediary term accounts for the
    $\Lambda t$ factor, the construction of $J^{iab}$ is responsable for the $dp$ term.
\end{itemize}

\section{Small fluctuations}
\label{app:small_fluct}
We want to calculate here the value of the inferred couplings under the MF approximation in the
case in which the fluctuations of the process $N_t$ are small.
Accordingly, we define the quantities
\begin{eqnarray}
    \delta k^{ia} & = & k^{ia} - \bar\Lambda^a \\
    \delta J^{iab} & = & J^{iab} - \delta^{i_a i_b} (1-\delta^{a0}) \frac{\bar\Lambda^a}{\bar\Lambda^i} \nu^{ab} - \frac{\bar\Lambda^a\bar\Lambda^b}{\bar\Lambda^i} \\
    & = & J^{iab} - J_0^{iab}\, \label{eq:def_j0},
\end{eqnarray}
which are expected to be small if the condition {\bf (MFH)} holds (explicit bounds on the rest terms are available in App.~\ref{app:cumulant_expansion}). In above equation we have also chosen the notation $\nu^{ab} = \int_0^\infty \dd t \, g^{a}(t) g^{b}(t)$, that we will employ extensively in the following.
Then it is possible to perturbatively invert the matrix $\bm{\mathsf{J}}$ by using the formula
\begin{equation}
    \bm{\mathsf{C}}^i = \bm{\mathsf{C}}^i_0 \sum_{n=0}^\infty (-\bm{\mathsf{\delta J}}^i{\bm{\mathsf{C}}_0^i})^n
\end{equation}
where we have defined the collection $\{ \bm{\mathsf C}^i_0 \}_{i=1}^d$ with elements
\begin{equation}
    {\bm{\mathsf C}^i_0 } = \bar\Lambda^i \left[
    \begin{array}{cc}
        1 + \sum_{j=1}^d (\bar\Lambda^j)^2 \sum_{q,q'=0}^p Z^{j}_{q q'} & \{\bm y^{j}\}_{1\leq j \leq d}^\top \\
        \, \{\bm y^{j}\}_{1\leq j \leq d}  & \left\{\delta^{jk} \bm{\mathsf{Z}}^{j} \right\}_{1\leq j,k \leq d} \\
    \end{array}
    \right] \, ,
\end{equation}
and introduced a collection of $d$ vectors $\{\bm{y}^{i}\}_{1\leq i \leq d}$ of dimension $p$ together with a collection of $d$ matrices of size $p\times p$ denoted as $\{\bm{\mathsf{Z}}^{i}\}_{1\leq i \leq d}$
\begin{eqnarray}
    \bm y^{i} &=& [-\bar\Lambda^i \sum_{q'=0}^p Z^{i}_{q q'}]_{1\leq q\leq p}  \\
    \bm{\mathsf Z}^{i} &=& [\bar\Lambda^i \nu^{ii}_{q q'}]^{-1}_{1\leq q,q' \leq p} \, ,
\end{eqnarray}
In above equation we have recovered the original notation of Sec.~\ref{sec:hawkes} which identifies kernel basis functions with subscripts, while the symbol $^{-1}$ above indicates the matrix inversion with respect to the $p\times p$ block.
One can easily verify by using above equation and Eq.~(\ref{eq:def_j0}) defining $\bm{\mathsf{J}}_0$ and $\bm{\mathsf{\delta J}}$ that
\begin{eqnarray}
    && \textstyle \bm{\mathsf{C}}^i_0 \, \bm{\mathsf{J}}^i_0 = \id \\
    && \textstyle \bm{\mathsf{C}}^i_0 \bm{\bar\Lambda} = \bm{\Sigma}^i \\
    && \textstyle \bm{\mathsf{\delta J}}^i \bm{\mathsf{C}}^i_0 \bm{\bar\Lambda} = \bar\Lambda^i \bm{\delta k}^{i} \, .
\end{eqnarray}
with $\bm\Sigma^i = [\Lambda^i \delta^{a0}]_{0\leq a \leq dp}$.
Thanks to these last two equations one can prove that
\begin{eqnarray}
    \bm\theta_{MF}^i &=& \bm{\mathsf{C}}^i (2\bm k^i - \bm h) \\
    \label{eq:mf_theta_simpler}
    &=& \bm{\Sigma}^i + \bm{\mathsf{C}}^i (\bm{\delta k}^i - \bm{\delta h})
\end{eqnarray}
\emph{at any order in} $\bm{\delta k}$! Together with the explicit expression for $\bm{\mathsf{C}}_0^i$, this is the main result of this appendix, that allows us to conclude that:
\begin{enumerate}
    \item The values of the interaction parameters $\bm\alpha$ are proportional to the fluctuations of $\bm{\delta k}^i$ and $\bm{\delta h}$ around their average values.
    \item When the fluctuations are very small, the MF approximation explains
    all the events by forcing the exogenous intensities $\bm\mu$ to be equal to the empirical ones
    $\bm{\bar\Lambda}$.
    \item One could be tempted to expand Eq.~(\ref{eq:mf_theta_simpler}) as
    \begin{equation}
        \bm\theta_{MF}^i = 
        \bm\Sigma^i
        + \bm{\mathsf{C}}_0^i (\bm{\delta k}^i - \bm{\delta h})
        - \bm{\mathsf{C}}_0^i \bm{\mathsf{\delta J}}^i \bm{\mathsf{C}}_0^i (\bm{\delta k}^i - \bm{\delta h})
        + \dots
    \end{equation}
    This is not always appropriate, as the truncation can lead to large errors even when
    the term $\bm{\mathsf{\delta J}}^i$ is extremely small. 
\end{enumerate}
In order to understand this last point, let us note that if one could take the limit of small $\bm{\mathsf{\delta J}}^i$ independently of the number of summands $dp$ in the tensor compositions, the approximation would be correct (this is appropriate in the regime of \emph{weak interactions} described in Sec.~\ref{sec:simulation}). On the other hand if the fluctuations are suppressed by the system size $d$, it is important to control the scaling in $d$ of $\bm{\mathsf{\delta J}}^i$, as one typically expects $\bm{\mathsf{\delta J}}^i \sim O(d^{-1})$, so that $\bm{\mathsf{\delta J}}^i \, \bm{\mathsf{C}}_0^i$ can be estimated to be of order 1 (this should be the case of \emph{well-mixed interactions} mentioned in Sec.~\ref{sec:simulation}) and the truncation leads to large errors.

\section{MLE versus MF solution} 
\label{app:mle_versus_mf_solution}

In this appendix we want to construct a bound on the error induced by the mean-field approximation, so to be able
to compare it with the typical value of the couplings and the statistical error due to the finite value of $T$.
In particular, we are interested in controlling the distance between the exact saddle-point solution
$\bm{\theta}^i_{MLE}$ and the mean-field one $\bm{\theta}^i_{MF}$ given by Eq.~(\ref{eq:estimator_theta}), that are given by
\begin{eqnarray}
    \bm{\theta}^i_{MLE} &=& \textrm{argmin}_{\bm\theta^i} \, \mathcal{L}(\bm N_t,\bm\theta)\\
    \bm{\theta}^i_{MF} &=&  \bm{\mathsf{C}}^i (2\bm{k}^i-\bm{h})
\end{eqnarray}
The idea is to bound their difference
\begin{equation}
    \bm{\delta\theta} = \bm{\theta}_{MLE} - \bm{\theta}_{MF} \, .
\end{equation}
where $\mathcal{L}(\bm N_t,\bm\theta)= -\log\prob(\bm N_t | \bm\theta)$ is the negative log-likelihood function equal to
\begin{equation}
    \mathcal{L}(\bm N_t,\bm\theta)= \bm{h}^\top\bm{\theta}  - \sum_{m=1}^n \log \lambda^{u_m}_{t_m} \, .
\end{equation}
Minimizing it with respect to $\theta^{ia}$ produces the relation:
\begin{eqnarray*}
    0 & = & h^{ia} - \frac{1}{T}\sum_{m=1}^n \frac{\partial \lambda^{u_m}_{t_m}}{\partial \theta^{ia}} \frac{1}{\lambda^{u_m}_{t_m}} \\
      & = & h^{ia} - \frac{1}{T}\sum_{m=1}^n \frac{\partial \lambda^{u_m}_{t_m}}{\partial \theta^{ia}}
      \left(
        \frac{1}{\bar\Lambda^{u_m}} - \frac{\lambda^{u_m}_{t_m}-\bar\Lambda^{u_m}}{(\bar\Lambda^{u_m})^2} + \frac{(\xi^{u_m}_{t_m}-\bar\Lambda^{u_m})^2}{(\bar\Lambda^{u_m})^3}
      \right) \\
      & = & h^{ia} - 2 k^{ia} + \sum_{b\in\mathcal{A}^i} J^{iab} \theta^{ib}_{MLE} - \frac{1}{T} \sum_{m=1}^n \frac{\partial \lambda^{u_m}_{t_m}}{\partial \theta^{ia}} \frac{(\xi^{u_m}_{t_m}-\bar\Lambda^{u_m})^2}{(\bar\Lambda^{u_m})^3}
\end{eqnarray*}
where in the secion line we have expanded $\lambda^{u_m}_{t_m}$ around $\bar\Lambda^{u_m}$, and we defined the rest of the Taylor expansion $\xi^i_t$, satisfying $| \xi^i_t / \bar\Lambda^i -1 | \leq | \lambda^i_t / \bar\Lambda^i -1 |$.

One can make some progress by first substituting the definition of $\bm{\delta\theta}^i$ into above equation, and then using the definition of $\bm{\theta}_{MF}^i$ in order to get:
\begin{equation}
    \label{eq:delta_theta}
    \bm{\delta\theta}^i = \bm{\mathsf C}^i {\bm v}^i \, ,
\end{equation}
where
\begin{eqnarray*}
    v^{ia}
    &=&
        \frac{1}{T} \sum_{m=1}^n \frac{\partial \lambda^{u_m}_{t_m}}{\partial \theta^{ia}}
        \frac{(\xi^{u_m}_{t_m}-\bar\Lambda^{u_m})^2}{(\bar\Lambda^{u_m})^3} \\
    &=&
        \frac{1}{N^i_T} \int \dd N_t^i \dd N_{t'}^a \, g^{a}(t-t') (\xi^i_t / \bar\Lambda^i -1)^2
\end{eqnarray*}
This allows us to construct an estimate for $\bm v$ in order to obtain a bound on $\bm{\delta\theta}$.

\paragraph{Expansion of $\bm{\delta\theta}$.}
One can first use a bound on the spectral norm of the $\bm{\mathsf{C}}^i$ matrices in order to constrain $\bm{\delta\theta}$. More precisely,
\begin{equation*}
    ||\bm{\delta\theta}^i|| \leq ||\bm{\mathsf{C}}^i|| ||\bm v^i||
\end{equation*}
with
\begin{eqnarray*}
        |v^{ia}| = v^{ia}
    &=&
        \frac{1}{N^i_T} \int_{t\neq t'} \dd N_t^i \dd N_{t'}^a \, g^{a}(t-t') (\xi^i_t / \bar\Lambda^i -1)^2 \\
    &\leq&
        \frac{1}{N^i_T} \int_{t\neq t'} \dd N_t^i \dd N_{t'}^a \, g^{a}(t-t') (\lambda^i_t / \bar\Lambda^i -1)^2 \\
    &=&
        k^{ia} - 2 \sum_{b=0}^{dp} J^{iab} \theta_{MLE}^{ib} +
        \sum_{b,c=0}^{dp} \theta_{MLE}^{ib}\theta_{MLE}^{ic} Q^{iabc} \, ,
\end{eqnarray*}
where we have used the relation
\begin{equation}
    \lambda^i_t = \sum_{a=0}^{dp} \theta_{MLE}^{ia} \int \dd N_{t'}^a \, g^{a}(t-t') \, .
\end{equation}
calculated at the saddle-point, and introduced a further auxiliary function
\begin{equation}
    Q^{iabc} = \frac{T^2}{(N^i_T)^3} \int_{t\neq t',t''.t'''} \dd N_t^i \dd N_{t'}^a \dd N_{t''}^b \dd N_{t'''}^c g^{a}(t-t') g^{b}(t-t'') g^{c}(t-t''') \, .
\end{equation}
This allows to rewrite more compactly
\begin{equation}
    \label{eq:v_bound}
    |v^{ia}| \leq \sum_{0\leq b,c \leq dp} (\theta_{MLE}^{ib}-\Sigma^{ib})Q^{iabc}(\theta_{MLE}^{ic}-\Sigma^{ic}) \, ,
\end{equation}
where $\bm\Sigma^i = [\bar\Lambda^i \delta^{a0}]_{0\leq a \leq dp}$ has been introducted in App.~\ref{app:small_fluct}. The next step will be to perform a cumulant expansion of $Q^{iabc}$ so to take into account the different contributions to Eq.~(\ref{eq:v_bound}).
\\
Before starting the inspection of the various cumulants of $Q^{iabc}$ one-by-one,
let us expand first the higher order differentials
$ \int_{t,t'} \dd N^a_t \dd N^b_{t'} = \int_{t\neq t'} \dd N^a_t \dd N^b_{t'} + \int_{t}\dd N^a_{t} \delta^{i_a i_b} (1 - \delta^{a0})$ by writing
\begin{eqnarray*}
    \label{eq:q_expansion}
        Q^{iabc}
    &=&
        \frac{T^2}{(N^i_T)^3} \int_{t\neq t'\neq t''\neq t'''} \dd N_t^i \dd N_{t'}^a \dd N_{t''}^b \dd N_{t'''}^c g^{a}(t-t') g^{b}(t-t'') g^{c}(t-t''') \\
    &+& \delta^{i_b i_c} (1-\delta^{b0})
        \frac{T^2}{(N^i_T)^3} \int_{t\neq t'\neq t''} \dd N_t^i \dd N_{t'}^a \dd N_{t''}^b g^{a}(t-t') g^{b}(t-t'') g^{c}(t-t'') \\
    &+& \delta^{i_a i_b} (1-\delta^{a0})
        \frac{T^2}{(N^i_T)^3} \int_{t\neq t'\neq t'''} \dd N_t^i \dd N_{t'}^a \dd N_{t'''}^c g^{a}(t-t') g^{b}(t-t') g^{c}(t-t''') \\
    &+& \delta^{i_a i_c} (1-\delta^{a0})
        \frac{T^2}{(N^i_T)^3} \int_{t\neq t'\neq t''} \dd N_t^i \dd N_{t'}^a \dd N_{t''}^b g^{a}(t-t') g^{b}(t-t'') g^{c}(t-t') \\
    &+&
        \delta^{i_a i_b}\delta^{i_a i_c} (1-\delta^{a0})
        \frac{T^2}{(N^i_T)^3} \int_{t\neq t'} \dd N_t^i \dd N_{t'}^a g^{a}(t-t') g^{b}(t-t') g^{c}(t-t')
\end{eqnarray*}
\paragraph{First order contribution}
At first order in the cumulants of $N_t$, obtained by substituting $\dd N_t = \bar\Lambda \dd t $, we have
\begin{eqnarray*}
        Q^{iabc}_{(1)}
    &=&
        \frac{\bar\Lambda^a }{(\bar\Lambda^i)^2} \left(\bar\Lambda^b\bar\Lambda^c +  \delta^{i_b i_c} \bar\Lambda^b (1-\delta^{b0}) \nu^{bc}  \right. \\
    &+& \left. \delta^{i_a i_b}(1-\delta^{a0})\bar\Lambda^c\nu^{ab} + \delta^{i_a i_c}(1-\delta^{a0})\bar\Lambda^b\nu^{ac} + \delta^{i_a i_b}\delta^{i_a i_c}(1-\delta^{a0}) \rho^{abc} \right) \, ,
\end{eqnarray*}
where $\rho^{abc} = \int_0^\infty \dd t \, g^{a}(t)g^{b}(t)g^{c}(t)$.
By using the properties of $\bm\Sigma$ proved in App.~\ref{app:small_fluct}, we can check that
\begin{eqnarray*}
        |v_{(1)}^{ia}|
    &=&
        \frac{\bar\Lambda^a }{(\bar\Lambda^i)^2} \bigg(
        (\delta\bar\Lambda^{i})^2 \\
    &+&
        \sum_{b,c = 0}^{dp}
        \theta^{ib}_{MLE} \theta^{ic}_{MLE}
        \bar\Lambda^b \nu^{bc} \delta^{i_b i_c} \\
    &+&
        2 (1-\delta^{a0}) \delta\bar\Lambda^i \sum_{b=0}^{dp} \nu^{ab} \delta^{i_a i_b} \theta^{ib}_{MLE} \\
    &+&
        \sum_{b,c = 0}^{dp}
        (1-\delta^{a0}) \rho^{abc}
        \theta^{ib}_{MLE} \theta^{ic}_{MLE} \delta^{i_a i_b} \delta^{i_a i_c}
        \bigg) 
\end{eqnarray*}
where we have defined
\begin{equation}
    \delta \bar\Lambda^i = \bm\theta^{i\top}_{MLE} \bm{\bar\Lambda} - \bar\Lambda^i
\end{equation}
which is expected to go to zero with $T$ independently of the mean-field approximation.
The leading order term is the one associated with the second line of Eq.~(\ref{eq:q_expansion}), as
\begin{itemize}
    \item The first and the third line are associated with terms that go to zero with $T$ due to the presence of $\delta\bar\Lambda^i$.
    \item The fourth line term is bound by a quantity the order of $ p^2 \, |\theta_{MLE}^{\max}|^2$, and hence is expected to be of the order of $p^2 ||c_{\max}||_1^2$ -- as one can show by using the bounds proven in App.~\ref{app:cumulant_expansion} -- where the functions $c^{ij}(t)$ are the empirical connected cross-correlation functions, which are expected to be of order $d^{-1}$.
    \item The second line term is instead of the order of $d p^2||c_{\max}||_1^2$
\end{itemize}
Hence, if $||c_{\max}||_1 \sim d^{-1}$, as one expects in the large dimensional setting (see e.g. App.~\ref{sec:homogeneous_system} which proves this scaling in the homogeneous case), that is of the order of $d^{-1}$.

\paragraph{Higher order contributions}

At the second order in the cumulants (i.e., terms containing the first power of the cross-correlation function $\bm{c}(t)$), all terms but one contain contributions smaller than $\sim d^{-1}$. In fact, all terms but one do either
\begin{itemize}
    \item contain one extra power of $||c_{\max}||_1$,
    \item or they are proportional to $\delta \bar\Lambda^i$ .
\end{itemize}
The only term which can contribute at order $d^{-1}$ is the one associated with the first line of Eq.~(\ref{eq:q_expansion}). Such term induces a contribution
\begin{eqnarray*}
        |v_{(2)}^{ia}|
    &=&
        \frac{\bar\Lambda^a }{(\bar\Lambda^i)^2} \sum_{b,c=0}^{dp}
        \theta^{ib}_{MLE} \theta^{ic}_{MLE}
        \int \dd t \dd t' c^{bc}(t-t') g^{b}(t)g^{c}(t') \\
    &+& (\textrm{higher order})
\end{eqnarray*}
Terms associated with cumulants of order larger than the second do not contribute at leading order. Hence, the dominant contribution is of the order of $d^{-1}$.
Putting this information together, we obtain
\begin{eqnarray*}
        |v^{ia}|
    &\leq& \left| 
        \frac{\bar\Lambda^a }{(\bar\Lambda^i)^2} \bigg[ \sum_{b,c=0}^{dp}
            \theta^{ib}_{MLE} \theta^{ic}_{MLE}
            \int \dd t \dd t' \left( c^{bc}(t-t') + \bar\Lambda^b \delta^{i_b i_c} \delta(t-t') \right) g^{b}(t)g^{c}(t')
        \bigg] \right| \\
    &+&
        (\textrm{higher order}) \, .
\end{eqnarray*}
that can be more compactly written as
\begin{equation*}
    \label{eq:variance_bound}
        |v^{ia}|
    \leq \left| 
        \frac{\bar\Lambda^a }{(\bar\Lambda^i)^2} \bar \var(\lambda^i_t)
        \right| + (\textrm{higher order})
\end{equation*}
where the variance corresponds to the \emph{empirical} one computed under the saddle-point parameters $\bm{\theta}_{MLE}$.
The term in parenthesis corresponds exactly to the fluctuations of $\bm N_t$, hence {\bf the error induced by the mean-field approximation is bound by the variance of the intensity $\bm\lambda_t$} as anticipated in the main text. We finally get:
\begin{equation}
    \label{eq:bound_error}
    || \bm{\delta\theta^i} || \leq ||\bm{\mathsf C}^i|| ||\bm{\bar\Lambda}||
    \frac{\bar\var(\lambda^i_t)}{(\bar\Lambda^i)^2} +
        (\textrm{higher order})
\end{equation}
that is the formula used in Sec.~\ref{sec:3.3}.


\section{Cumulant expansion} 
\label{app:cumulant_expansion}
In this appendix we will be interesting in providing \emph{a priori} bounds on the value of the inferred parameters $\bm\theta_{MF}^i$ and on their associated errors $\bm{\delta\theta}^i$. We will be able to relate with quantities with the empirical cumulants of the process $\bm N_t$, thus allowing to obtain bounds on the error that do not require assumptions about the underlying model.

We start by recalling that after neglecting border effects (i.e., in the limit of large $T$), one has:
\begin{eqnarray}
    h^{ia}  & = & \bar\Lambda^a \\
    \label{eq:k_cum_exp}
    k^{ia}  & = & \bar\Lambda^a + \frac{1}{\bar\Lambda^i}\int_0^\infty \dd t \, c^{ia}(t) g^{a}(t) \\
    J^{iab} & = & \delta^{i_a i_b} (1-\delta^{a0})\left(
        \frac{\bar\Lambda^a}{\bar\Lambda^i} \int_0^\infty \dd t \, g^{a}(t) g^{b}(t)
        +
        \frac{1}{(\bar\Lambda^i)^2} \int_0^\infty \dd t \, c^{ia}(t) g^{a}(t) g^{b}(t)
    \right) \nonumber \\
    &+& \frac{\bar\Lambda^a \bar\Lambda^b}{\bar\Lambda^i}
    + \frac{1}{\bar\Lambda^i}\int_0^\infty \dd t \dd t' \, c^{ab}(t-t') g^{a}(t) g^{b}(t') \\
    &+&
      \frac{\bar\Lambda^a}{(\bar\Lambda^i)^2}\int_0^\infty \dd t \, c^{ib}(t) g^{b}(t)
    +
    \frac{\bar\Lambda^b}{(\bar\Lambda^i)^2}\int_0^\infty \dd t \, c^{ia}(t) g^{a}(t) \nonumber \\
    &+&  \frac{1}{(\bar\Lambda^i)^2} \int_0^\infty \dd t \dd t' \, K^{iab}(t,t') g^{a}(t) g^{b}(t') \nonumber \,
\end{eqnarray}
where the vectors ${\bm c}^i(t)$ and the matrices $\bm{\mathsf{K}}^i(t,t')$ correspond respectively to the
second and the third empirical cumulant of the process $N_t$, which for $t,t' > 0$ and $a,b >0$ are defined as
\begin{eqnarray}
    c^{ij}(t)  & = & \frac{1}{T} \sum_{m' < m} \delta^{i u_m} \delta^{j  u_{m'}} \delta(t-t_m + t_{m'}) - \bar\Lambda^i\bar\Lambda^j \\
    K^{ijk}(t,t')  & = & \frac{1}{T} \sum_{m'\neq m'' < m}
    \delta^{i u_m} \delta^{j  u_{m'}} \delta^{k  u_{m''}} \delta(t-t_m + t_{m'}) \delta(t'-t_m + t_{m''}) \nonumber \\
    &-& \bar\Lambda^i\bar\Lambda^j\bar\Lambda^k - \bar\Lambda^i c^{jk}(t-t') - \bar\Lambda^j c^{ik}(t') - \bar\Lambda^k c^{ij}(t) \,
\end{eqnarray}
while one obviously has $c^{i0}(t) = K^{i0a}(t,t') = K^{ia0}(t,t') = 0 $. As in the last appendix, we write $\nu^{ab} = \int_0^\infty \dd t \, g^{a}(t) g^{b}(t)$, and we will also employ
the quantities
\begin{eqnarray}
    \delta k^{ia} & = & k^{ia} - \bar\Lambda^a \\
    \delta J^{iab} & = & J^{iab} - \delta^{i_a i_b} (1-\delta^{a0}) \frac{\bar\Lambda^a}{\bar\Lambda^i} \nu^{ab} - \frac{\bar\Lambda^a\bar\Lambda^b}{\bar\Lambda^i} \\
    & = & J^{iab} - J_0^{iab}\,,
\end{eqnarray}
which are proportional to the empirical cumulants. The quantities $\bm{\delta k}^i$ and $\bm{\mathsf{\delta J}}^i$ are in fact bound by
\begin{eqnarray}
    | \delta k^{ia} | & \leq &  \frac{g_{\max}||c_{\max}||_1}{\bar\Lambda_{\min}}\\
    | \delta J^{iab}| & \leq & 
    \delta^{i_a i_b} (1-\delta^{a0}) \left( \frac{g^2_{\max}||c_{\max}||_1}{\bar\Lambda^2_{\min}} \right) \\
    & + & 2 \left( \frac{g_{\max}||c_{\max}||_1}{\bar\Lambda_{\min}} \right)
    + \frac{g^2_{\max}||K_{\max}||_1}{\bar\Lambda^2_{\min}}
\end{eqnarray}
where $\bar\Lambda_{\min} = \min_{1\leq i\leq d} \bar\Lambda^i$, $g_{\max} = \max_{a,t} g^{a}(t)$ and
\begin{eqnarray}
    ||c_{\max} ||_1 & = & \max_{ij} \int_0^\infty \dd t \, |c^{ij}(t)| \\
    ||K_{\max} ||_1 & = & \max_{ijk} \int_0^\infty \dd t \dd t' \, | K^{ijk}(t,t') | \, .
\end{eqnarray}
By particularizing these inequalities to Eq.~(\ref{eq:variance_bound}) one can obtain a 
bound on the mean-field error that is of the form
\begin{eqnarray*}
        |v^{ia}|
    &\leq& 
        \frac{\bar\Lambda^a }{(\bar\Lambda^i)^2} \left(|| c_{\max} ||_1 g_{\max} \left(\sum_{b=0}^{dp} \theta_{MLE}^{ib}\right)^2 + \nu_{\max} \bar\Lambda_{\max} \sum_{b,c=0}^{dp} \theta_{MLE}^{ib}\theta_{MLE}^{ic}\delta^{i_b i_c} \right) \\
    &\leq&
        \frac{|| c_{\max} ||_1 g_{\max}\bar\Lambda_{\max}}{\bar\Lambda_{\min}^2} \left(\sum_{b=0}^{dp} \theta_{MLE}^{ib}\right)^2
    +
        \frac{\nu_{\max} \bar\Lambda_{\max}^2}{\bar\Lambda_{\min}^2}
        \sum_{b,c=0}^{dp} \theta_{MLE}^{ib}\theta_{MLE}^{ic}\delta^{i_b i_c}
\end{eqnarray*}
By constructing the norm of the error, and comparing them with the bound on the inferred couplings Eq.~(\ref{eq:mf_theta_simpler}), yields the expressions
\begin{eqnarray*}
    || \bm\theta^i_{MF}|| &\leq& \sqrt{dp} ||\bm{\mathsf{C}}^i|| \left( \frac{g_{\max}||c_{\max}||_1}{\bar\Lambda_{\min}} \right) \\
    || \bm{\delta\theta}^i|| &\leq& \sqrt{dp} ||\bm{\mathsf{C}}^i|| \left(
        \frac{|| c_{\max} ||_1 g_{\max}\bar\Lambda_{\max}}{\bar\Lambda_{\min}^2} \left(\sum_{b=0}^{dp} \theta_{MLE}^{ib}\right)^2
    +
        \frac{\nu_{\max} \bar\Lambda_{\max}^2}{\bar\Lambda_{\min}^2}
        \sum_{b,c=0}^{dp} \theta_{MLE}^{ib}\theta_{MLE}^{ic}\delta^{i_b i_c}
    \right) \, .
\end{eqnarray*}


\bibliographystyle{unsrt}
\bibliography{mf_hawkes}

\end{document}